\documentclass[manuscript,balance=false]{acmart}
\usepackage{caption}
\usepackage{subcaption}
\usepackage{makecell}
\usepackage{multirow}
\usepackage{amsmath}
\usepackage[linesnumbered,ruled,vlined]{algorithm2e}
\usepackage{arydshln}

\acmConference[Woodstock '18]{Woodstock '18: ACM Symposium on Neural Gaze Detection}{June 03--05, 2018}{Woodstock, NY}

\acmBooktitle{}

\begin{document}

\settopmatter{printacmref=false}
\setcopyright{none}
\renewcommand
\footnotetextcopyrightpermission[1]{}
\pagestyle{plain}

\title{Analysing Fairness of Privacy-Utility Mobility Models}

\author{Yuting Zhan, Hamed Haddadi }
\affiliation{Imperial College London\country{UK}}

\author{Afra Mashhadi }
\affiliation{University of Washington\country{USA}}

\renewcommand{\shortauthors}{Zhan et al.}


\begin{abstract}
 
Preserving the individuals' privacy in sharing spatial-temporal datasets is critical to prevent re-identification attacks based on unique trajectories. Existing privacy techniques tend to propose ideal privacy-utility tradeoffs, however, largely ignore the fairness implications of mobility models and whether such techniques perform equally for different groups of users. The quantification between fairness and privacy-aware models is still unclear and there barely exists any defined sets of metrics for measuring fairness in the spatial-temporal context. In this work, we define a set of fairness metrics designed explicitly for human mobility, based on structural similarity and entropy of the trajectories. Under these definitions, we examine the fairness of two state-of-the-art privacy-preserving models that rely on GAN and representation learning to reduce the re-identification rate of users for data sharing. Our results show that while both models guarantee group fairness in terms of demographic parity, they violate individual fairness criteria, indicating that users with highly similar trajectories receive disparate privacy gain. We conclude that the tension between the re-identification task and individual fairness needs to be considered for future spatial-temporal data analysis and modelling to achieve a privacy-preserving fairness-aware setting.

\end{abstract}

 



\maketitle

\section{Introduction}
Understanding human mobility based on collected locations from mobile devices has become a fundamental part of urban and environmental planning in cities~\cite{luca2021survey}. These GPS traces enable the scientific community and policymakers to model citizens' daily mobility patterns (\emph{e.g.}, crowd-sensed car sharing, ride sharing, city bicycles sharing, and RFID-card-based public transportation, or build predictive algorithms to estimate people’s flows and community structure~\cite{ferreira2020deep}. However, location-based traces corresponding to human mobility, even at an aggregate level, have raised numerous privacy concerns~\cite{song2010limits,de2013unique}, mainly when the data contains sensitive and revealing insights about people’s identity, behaviour, associations, religion, and others~\cite{kamargianni2015feasibility}.

In the past decades, the research community has examined various ways of ensuring the privacy of mobility traces. Previous work, ranging from k-anonymity~\cite{aristodimou2016privacy}, differential privacy (\emph{i.e.,} DP)~\cite{saleheen2016msieve,xiao2015protecting}, to information-theoretic metrics~\cite{puttaswamy2012preserving,zhang2018online}, explore scientific guarantees that data subjects cannot be re-identified while the data remain practically useful.
More recently, \textit{privacy-utility trade-off} (\textit{PUT}) models based on machine learning or deep learning techniques that aim to optimize both privacy and utility (\emph{i.e.}, inference accuracy) have been studied and shown to be superior to the previous approaches~\cite{erdemir2020privacy}. 
These techniques can be summarized as representation learning~\cite{huang_variational_2019}, generative adversarial network (GAN)~\cite{rao2020lstm,ijcai2018-530}, reinforcement learning~\cite{erdemir2020privacy,erdemir2019privacy}, etc. 
In these works, researchers have shown that it is possible to design and implement frameworks that enhance the privacy protection of individual trajectories without a significant reduction of  the trace's utility.

A dimension that has been vastly overlooked is whether privacy-preserving algorithms work equally for all users or whether they could lead to unexpected consequences of protecting the privacy of only a group of people. Indeed, as recent evidence from the broader machine learning domain has shown, the systematic discrimination in making decisions against different groups has been shifted from people to autonomous algorithms~\cite{kasy2021fairness,heidari2019moral}. In many applications, discrimination may be defined by different protected attributes, such as race, gender, ethnicity, and religion, that directly prevent favourable outcomes for a minority group in societal resource allocation, education equality, employment opportunity, etc~\cite{sattigeri2019fairness}. Similarly, in the context of spatial-temporal data, mobility demand prediction algorithms have been shown to offer higher service quality to neighbourhoods with more white people~\cite{brown2018ridehail}. However, in such contexts, only a handful of recent studies exist that examine the fairness of location-based systems~\cite{yan2019fairst,yan2020fairness,ge2016racial}, with little consensuses on how fairness should be defined and measured for spatial-temporal applications. 

In this work, we aim to measure and evaluate the fairness of the location privacy-preserving algorithms applied to mobility traces. We seek to answer the research question as to {\em whether the outcome of the PUT models satisfies fairness}. 
Extended from the notion of fairness in broader machine learning literature, fairness in location privacy-preserving mechanisms could also be concluded in two categories: \emph{individual fairness} and \emph{group fairness}. \emph{Individual fairness} ensures that similar users receive similar outcomes with respect to the specific privacy-aware inference tasks~\cite{binns2020apparent}. 
That is, whether these privacy-aware models preserve the privacy and service quality of similar users equally.  
In order to do so, we first posit a set of similarity metrics to mathematically denote a notion of user similarity grounded on the human mobility literature~\cite{friedler2021possibility}, in terms of both the structural similarity of their heatmap images and the entropy of their trajectories. 
On the other hand, \emph{group fairness} ensures the independence between the model outcome and a sensitive attribute (i.e., gender, age, ethnicity, etc) of interest.
That is, it ensures equal privacy gain and utility loss over multiple groups.

We examine two machine learning-based privacy-preserving approaches (i.e., TrajGAN~\cite{rao2020lstm} and Mo-PAE~\cite{zhan2022privacyaware}), compared to the original inference tasks that optimize only for privacy or for utility. We evaluate their fairness on two real-life mobility datasets: Geolife~\cite{zheng_geolife_2010} and MDC~\cite{laurila2012mobile}.
Our results indicate that both TrajGAN and Mo-PAE do not guarantee \emph{individual fairness}; users with similar trajectories might receive different privacy gain outcomes where the \emph{individual fairness} criteria are violated in these location privacy-aware settings. 
More specifically, we observe that for the users with similar traces, even when the outcome of the prediction task is identical, the privacy gains amongst those users are highly different, leading to some users not advantaging from obfuscation as others do.
Different to the highlights of individual fairness, the results of \emph{group fairness} of privacy-aware models show that there is no demographic disparity in the privacy and prediction outcome. 
However, as we discuss, this observation highly reflects the socio-cultural settings where these traces have been collected and are less of a by-product of the privacy-preserving models. 
The contributions of our paper are as follows:
\begin{itemize}
    \item We theoretically denote the notion of \emph{similarity} for tackling the measurements of individual fairness of spatial-temporal datasets.
    \item We offer a set of \emph{individual fairness} metrics specifically defined based on mobility characteristics that can help the broader research community measure fairness for spatial-temporal applications.
    \item We examine the privacy-preserving algorithms in terms of both individual fairness and group fairness on two representative mobility datasets, and show their deficiencies in accounting for fairness can lead to undesired consequences. 
    \item We systematically discuss why individual fairness and group fairness are competing in the privacy-aware setting.
\end{itemize}

\section{Related work}\label{sec:related}

\subsection{Fairness in Machine Learning}
\label{sec:related_FML}
Literature on fairness in machine learning (Fair-ML) tends to focus on \textit{the absence of any prejudices or favoritism toward an individual or group based on their inherent or acquired characteristics}~\cite{mehrabi2021survey}. The majority of fairness research strives to avoid the decision made by automated systems skewed toward the advantaged groups or individuals.  In~\cite{friedler2021possibility}, authors proposed a framework for understanding different definitions of fairness through two views of the world: i) {\em we are all equal} ({\em WAE}, mostly ensuring the \emph{group fairness}), and ii) {\em what you see is what you get} ({\em WYSIWYG}, mostly ensuring the \emph{individual fairness}). 
The framework shows that the fairness definitions and their implementations correspond to different axiomatic beliefs about the world, described as two fundamentally incompatible worldviews. A single algorithm cannot satisfy either definition of fairness under both worldviews~\cite{friedler2021possibility}.

The most adopted metrics for fairness in machine learning are widely based on the \textit{WAE} assumption and denoted as \emph{group fairness}, which is also known as \emph{statistical parity} and \emph{demographic parity}~\cite{dwork2012fairness}.
These metrics aim to ensure that there is independence between the predicted outcome of a model and sensitive attributes of age, gender, and race. 
If variations of \emph{statistical parity} exist, Fair-ML will concentrate on relaxation of this measure by ensuring that groups from sensitive attributes and non-sensitive attributes meet the same misclassification rate (\emph{false negative rate}, also known as \emph{equalized odds}~\cite{hardt2016equality}), or \emph{equal true positive rate} (also known as \emph{equal opportunity}~\cite{hardt2016equality}). 

In the context of mobility data and its applications, such as equitable transportation, research attention also mainly devoted to group fairness. Transportation equity heavily employs statistical tests for equity analysis, which is appropriate for discovering unfairness~\cite{yan2020fairness}. The such metric is often defined based on census tract information, which offers an aggregate demographic characteristic of the residing population. \citet{yan2019fairst} defined fairness in terms of the region-based fairness gap and assesses the gap between mean per capita ride-sharing demand across groups over time. The two metrics differ from each other. One is based on a binary label associated with the majority of the sub-population (e.g., white), and the other is based on a continuous distribution of the demographic attributes. Similarly, \citet{he2022socially} proposed a  graph-based approach for integrating group-based (census) into e-scooter demand prediction.  Through the integration of an optimization regularizer, they showed that it is possible for their model to jointly learn the flow patterns and socio-economic factors, and returns socially-equitable flow predictions. Hosford et al.~\cite{hosford2018public} investigated the equity of access to bike sharing in multiple cities in Canada. Ge et al.~\cite{ge2016racial} studied racial and gender discrimination in the expanding transportation network companies. These handfuls of recent works all focus on group-based fairness metrics and collective methods (e.g., demand or flow prediction).

On the other hand, individual fairness claims that similar individuals should be treated similarly concerning the target task~\cite{dwork2012fairness}.
For example, in making hiring decisions, the algorithm has to possess perfect knowledge of comparing the ``qualification'' of two individuals. 
In most cases, the difficulty with individual fairness lies in the notion of measuring {\em similarity}. For example, \citet{yan2020fairness} used the population and employment density of each city area for achieving individual fairness in bike-sharing demand prediction. 
The difficulty, again, lies in the fact that there is often a lack of perfect knowledge to determine the \emph{similarity} in demand between two areas.
In broader spatial-temporal data and application, the definitions of mobility similarity are almost non-existence, so as individual fairness of spatial-temporal data. Although previous work in fairness literature~\cite{chang2021privacy} has examined the boundary of fairness and privacy, these works have been applied to low dimensional datasets (e.g., COMPAS) that differ greatly from complex mobility data of people. In this work, we offer a new perspective on how to measure individual fairness metrics defined based on the literature on mobility and examine its application in assessing the fairness of the privacy-preserving algorithms applied to mobility traces.

\subsection{Privacy Methods for Spatial-Temporal Data}

Large-scale human mobility data contain crucial insights into understanding human behaviour but are hard to share in non-aggregated form due to their highly sensitive nature. Decades of research on privacy examined various anonymous mechanisms on human trajectories~\cite{saleheen2016msieve,xiao2015protecting,aristodimou2016privacy}. 
A mobility privacy study conducted by De Montjoye et al~\cite{de2013unique} illustrates that four spatial-temporal points are enough to identify 95\% of the individuals in a certain granularity, demonstrating the necessity of the anonymous mechanism against the re-identification attack.
Previous work, ranging from k-anonymity~\cite{aristodimou2016privacy}, differential privacy~\cite{saleheen2016msieve,xiao2015protecting}, to information theoretic metrics~\cite{puttaswamy2012preserving,zhang2018online}, explore scientific guarantees that the subjects of the data cannot be re-identified while the data remain practically useful.
More recently, PUT models based on machine learning, which simultaneously aim to optimize for data privacy protection and utility, are emerging. In these lines of work, researchers have focused on the objective of training neural network models that optimize for reducing privacy leakage risk of individual trajectories while at the same time minimizing the depreciation in the mobility utility.
These models have been shown to be superior to differential privacy techniques. In this paper, we selected two machine learning-based PUT models based on two different strategies of GAN and Representation Learning, but both with promising  high performance in terms of both utility and privacy. These two PUT models mainly focus on temporal correlations in time-series data and aim to reduce the user re-identification risk (i.e., privacy) while minimizing the downgrade in the accuracy of mobility prediction task (i.e., utility). We  describe the details of these two privacy-aware spatial-temporal models:

\textbf{TrajGAN}~\cite{rao2020lstm}: it is an end-to-end deep learning model to generate synthetic data that preserves the real trajectory data's essential spatial, temporal, and thematic characteristics. Compared with other standard geo-masking methods, TrajGAN can better prevent users from being re-identified. 
TrajGAN claims to preserve essential spatial and temporal characteristics of the original data, verified through statistical analysis of the generated synthetic data distributions, which is in a line with the data utility assessment based on the mobility prediction task in our work. Hence, we train a TrajGAN-based PUT model to evaluate the mobility predictability and privacy protection of synthetic data generated by TrajGAN.

\textbf{Mo-PAE}~\cite{zhan2022privacyaware}: it is a \textbf{p}rivacy-preserving \textbf{a}dversarial feature \textbf{e}ncoder. In contrast to the TrajGAN that aims to generate synthetic data, Mo-PAE trains an encoder $Enc_L$ that forces the extracted representations \textit{f} to convey maximal information about data utility while minimizing private information about user identities via adversarial learning. 
It consists of a  multi-task adversarial network to learn an LSTM-based encoder $Enc_L$,  which can generate the optimized feature representations $f=Enc_L(X)$ via lowering the privacy disclosure risk of user identification information (i.e., privacy) and improving the mobility  prediction accuracy (i.e., utility) concurrently. 

\section{Fairness Definition and Metrics}\label{sec:problem}
In this section, we first define the mathematical representation of fairness in spatial-temporal applications before we incorporate it into our analysis.

\subsection{Formulation of the Problem}
In this work, we aim to measure and evaluate the fairness of the privacy-preserving algorithms applied to mobility traces. We seek to figure out whether these models equally preserve the user privacy and inference accuracy of similar users. We try to determine whether fairness metrics benefit from a privacy-preserving model simultaneously, laying a theoretical foundation for further research on the privacy-preserving fairness-aware mechanism for human mobility. Both individual- and group-based fairness are discussed.

We first introduce some basic notations and abbreviations utilized in this work: individuals are labelled as \textit{u}, if individuals $u_i$ and $u_j$ are similar, that is $u_i \backsim u_j$; sensitive or protected attributes are denoted as \textit{S}; raw data without sensitive attributes is denoted as $X$; $Y$ is the ground-truth labels for a specific inference task and $Y'$ is the predicted one, which is the variant that depends on $S$ and $X$. The true positive rate (i.e., TPR, recall, or sensitivity) is utilized to judge the performance of the multi-categorical classifiers, which refers to the proportion of who should be predicted accurately that received a positive result. 
TPR is also utilized in the inference tasks' quality of the examined models and is denoted as \textit{task accuracy}.

\subsection{Individual Fairness}
\label{section:IF}

Individual fairness~\cite{dwork2012fairness} states that individuals who are similar, with respect to a specific task, should be treated similarly (i.e., $P_{u_i} \backsim P_{u_j}$ when $u_i \backsim u_j$)~\cite{roemer2002equality}:
\begin{equation}
\begin{aligned}
    P(Y'|u_i, S, X) = P(Y'|u_j, S, X)
\end{aligned}
\label{equ:individual fairness}
\end{equation}

As we have mentioned in Section~\ref{sec:related_FML}, the difficulty with individual fairness lies in the notion of measuring $similarity$.
To measure individual fairness in the context of spatial-temporal data, we need two sets of definitions corresponding to i) the similarity between users' {\em trajectories} ($SIM_t$); and ii) the similarity of the {\em outcome} of the PUT models ($SIM_o$), as well as  their generalizability for different mobility datasets and PUT models. We define each next:

\subsubsection{Similarity of Trajectories}
\label{section:sim of Tra}
Grounded on the literature on mobility~\cite{song2010limits,IJCNN,ferreira2020deep}, we mathematically denote the notion of trajectory similarity ($SIM_t$) based on i) the \textit{structural similarity index} of mobility heatmap images; and ii) the {\em entropy} of trajectories. 

\begin{figure*}
     \centering
     \begin{subfigure}[b]{0.39\textwidth}
         \centering
         \includegraphics[width=\textwidth]{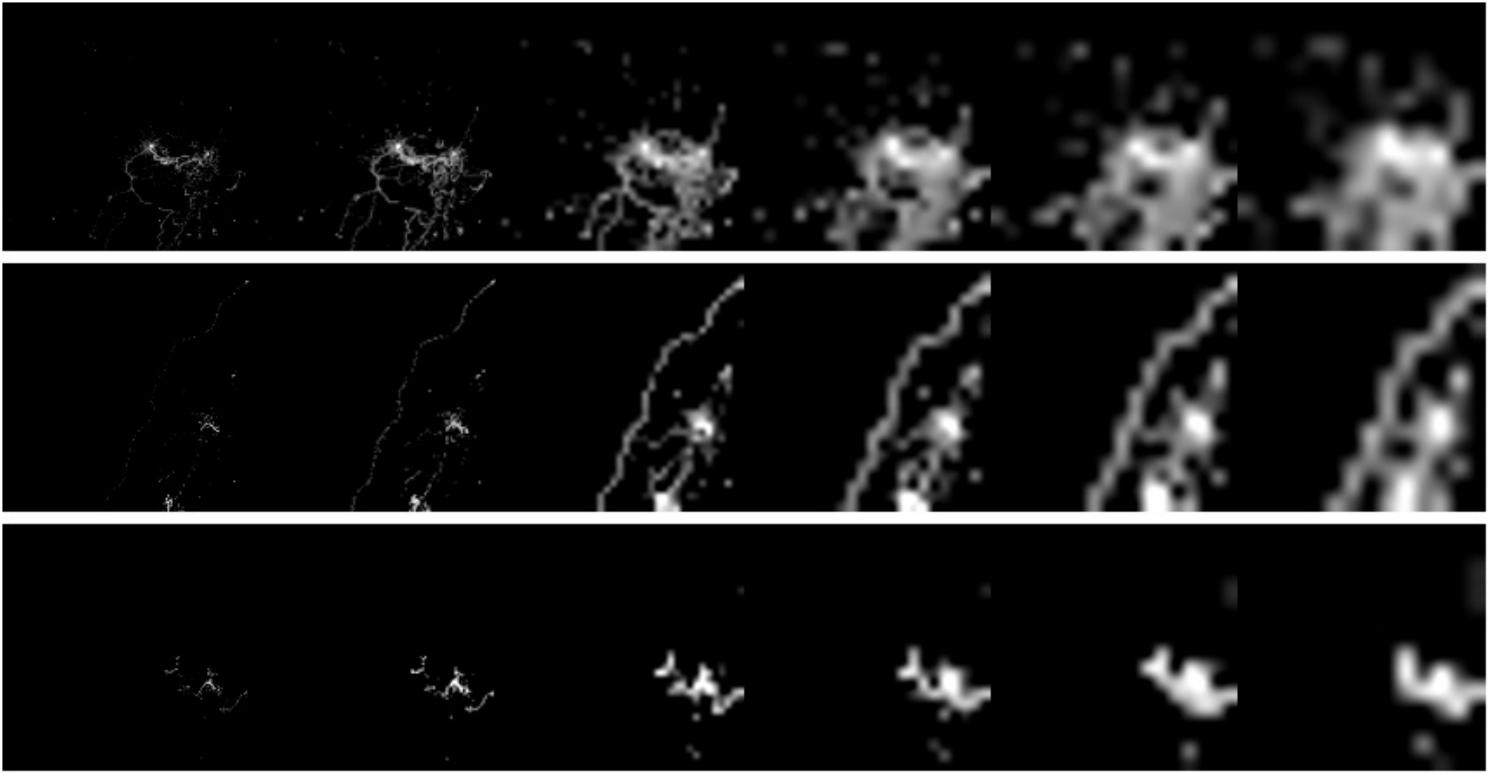}
         \caption{MDC}
         \label{fig:mdcsample}
      \end{subfigure}
      \begin{subfigure}[b]{0.37\textwidth}
         \centering
         \includegraphics[width=\textwidth]{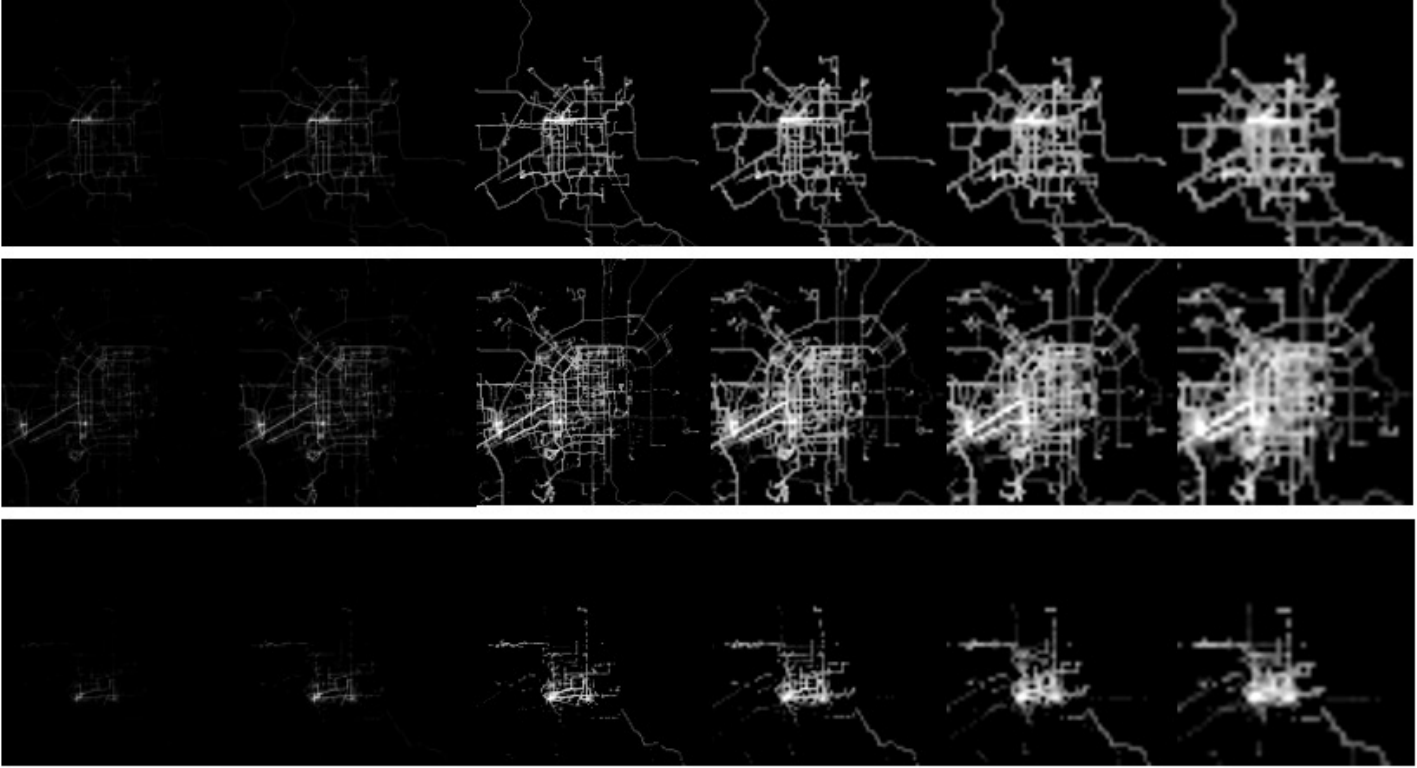}
         \caption{Geolife}
         \label{fig:geolifesample}
      \end{subfigure}
    \caption{Sample mobility heatmap images with various spatial granularities of MDC and Geolife. Three different trajectories are shown with different granularities (50 m, 100 m, 300 m, 500 m, 700 m, and 900 m). }
    \label{fig:heatmaps}
\end{figure*}

\textbf{Structural  Similarity Index Measure (SSIM)}: SSIM was initially designed to quantify image quality degradation caused by processing, such as data compression or losses in data transmission, which leverages the differences between the reference image and the processed image~\cite{wang2004image}.
To apply SSIM metrics in this work, we construct \emph{heatmap} images from the raw geo-located data with the methodology proposed by~\cite{ferreira2020deep}. Figure~\ref{fig:heatmaps} shows some sample heatmap images with spatial granularity coarsening from 50 meters to 900 meters by the left to right.
These heatmap images structurally represent mobility features extracted from mobility traces, which use pixel intensity to encode the {\em frequency} of the visit spent in a given area; hence, the brighter pixels denote the more frequently visited locations of the user. SSIM has been shown to be a well-suited metric to compute the image similarity of the  heatmap images apecifically when applied to mobility heatmap images~\cite{IJCNN,ferreira2020deep}. Unlike Mean Square Error, the SSIM metric has been shown not to be significantly impacted by the changes in luminosity and contrast.

In this work, we formulate the SSIM measure as the perceptual difference of two similar users' heatmap images, $H_i$ and $H_j$. See the Appendix for full definitions.
We then leverage the integrated heatmap image, which combines all user trajectories, to calculate the effective SSIM index ($SSIM_{eff}$) that indicates the overall trajectory similarity of users. The SSIMs between individual ($SSIM_{one}$) and integrated trajectory ($SSIM_{eff}$) are denoted by calculating the SSIM $maps$ (i.e., local values of the SSIM, $SSIM_{maps}=abs(SSIM_{eff}-SSIM_{one})$). $SSIM_{maps}$ is utilized to lower the impact of the unreached area, that is, only the swept area in the integrated heatmap image was selected for further analysis. Hence, the average SSIM value of the selected points is what we define as $SSIM_{eff}$. 
Additionally, as this metric relies on heatmap images, it is highly influenced by spatial granularity, where each pixel in the image corresponds to the spatial boundary of the data. Intuitively, in Figure~\ref{fig:heatmaps}, as the granularity coarsens, the trajectories become blurry and, thus, more similar. 
The impact of the spatial granularity on the SSIM index will discuss in Section~\ref{section:if_one}.

\begin{figure*}
     \centering
     \begin{subfigure}[b]{0.80\textwidth}
        \centering
        \includegraphics[width=\textwidth]{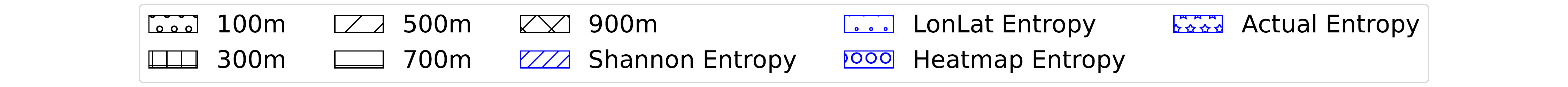}
        \label{fig:ssim_legend}
     \end{subfigure}
     \begin{subfigure}[b]{0.38\textwidth}
         \centering
         \includegraphics[width=\textwidth]{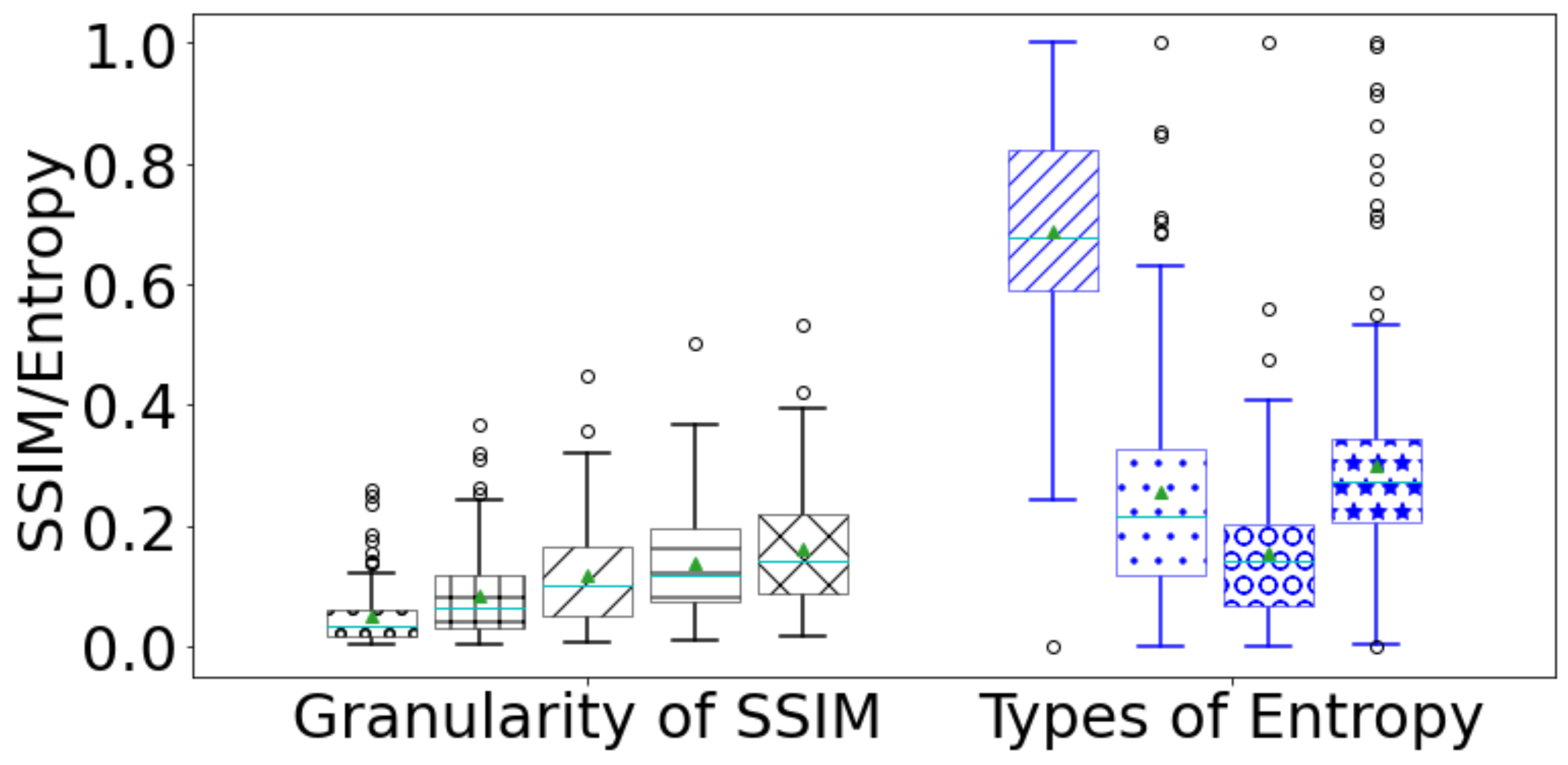}
         \caption{MDC}
         \label{fig:overallssim}
      \end{subfigure}
      \begin{subfigure}[b]{0.38\textwidth}
         \centering
         \includegraphics[width=\textwidth]{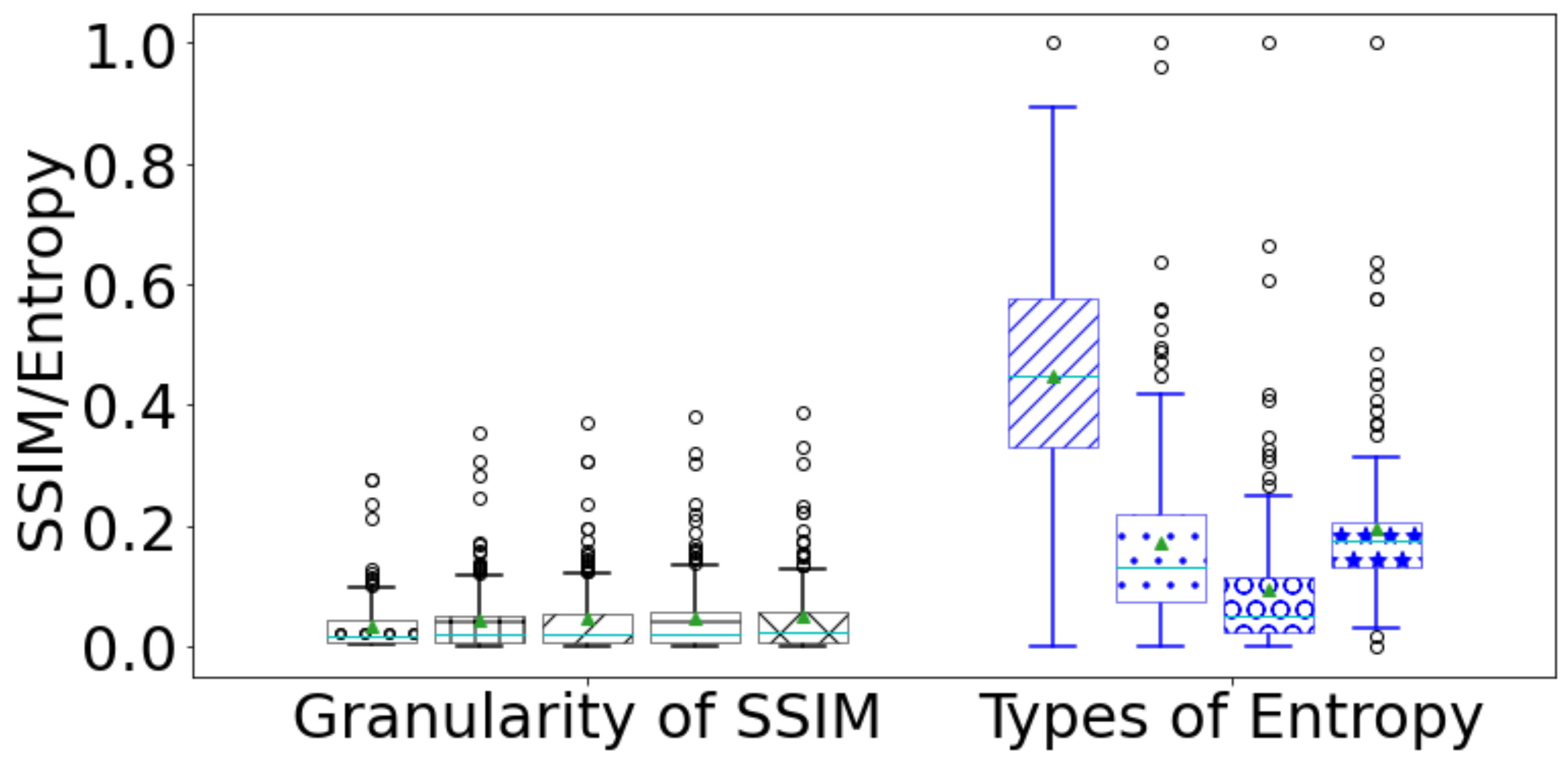}
         \caption{Geolife}
         \label{fig:overallssim}
      \end{subfigure}
    \caption{Overview of SSIM and entropy distribution of trajectories of MDC and Geolife datasets. Different granularities of SSIM are compared in a row, where the granularity are ranging from 100-meter to 900-meter. }
    \label{fig:ssimgran}
\end{figure*}

\textbf{Entropy of Trajectories (EOTs)}: Mobility literature defines the highest potential accuracy of predictability of any individual, termed as ``maximum predictability'' ($\Pi_{max}$)~\cite{lu2013approaching}. Maximum predictability is determined by the $entropy$ of a person's trajectory information (\emph{e.g.}, frequency, sequence of location visits, etc.). 
Hence, some similar characteristics of user spatial-temporal patterns are able to be captured by leveraging the entropy of trajectory. 
In this paper, we conclude and define four types of entropy to measure trajectory similarity for spatial-temporal applications, denoted as \emph{Shannon Entropy (SE), LonLat Entropy (LE), Heatmap Entropy (HE), Actual Entropy (AE)}. The integrated entropy of these four different types of entropy is denoted as EOTs. The details of them are as followed, and see the Appendix for full definitions.

i) {\em Shannon Entropy} (\textit{SE}): the entropy of probabilities of visited location distribution. 
Leveraging the common definition of Shannon entropy ($E_h$), a classic notion of data uncertainty, we first calculate $E_h$ of the trajectory to characterize visited location distribution and their probabilities.
A larger $E_h$ indicates greater disorder and consequently reduces the predictability of an individual's movements.

ii) {\em LonLat Entropy} (\textit{LE}): 
the entropy of the geo-located locations in a time-series format.
Considering the spatial-temporal pattern of the mobility data, the entropy of visited locations in terms of longitudes and latitudes are separately estimated by using the fuzzy entropy $E_f$. This entropy reflects the probability of a new sub-string and quantifies the irregularity or complexity of the time-series data.

iii) {\em Heatmap Entropy} (\textit{HE}): the entropy of the users' heatmap images. In contrast to the aforementioned entropy models, we define a two-dimensional entropy ($E_{2D}$) to quantify the irregularity (i.e., unpredictable dynamics) of the user's heatmap image. The entropy of trajectory heatmap images is calculated using the two-dimensional sample entropy method ($SampEn_{2D}$) \citep{silva2016two}. In a trajectory heatmap image, the features are extracted by accounting for the spatial distribution of pixels in different $m$-length square windows.

iv) {\em Actual Entropy} (\textit{AE}): the entropy of capturing entire spatial-temporal order present in user's mobility pattern.
To capture AE, \citet{song2010limits} proposed an actual entropy model using the Lempel-Ziv algorithm. Different to other types of entropy, AE depends not only on the frequency of visited locations but also on the order in which the nodes were visited and the time spent at each location~\cite{song2010limits}.
In this work, the given area is segmented using structured grids, where each grid is initialized as $0$. Then the visited locations and whether the person reached the cell previously are tracked. If the person visits an unreached cell, the location is marked as $1$, generating time-series binary data to characterize the trajectory. 

See the Appendix for full definitions and related equations of these four different entropies.

\subsubsection{Similarity of Users}
With the aforementioned definition of trajectory similarity, we mathematically  define the users with similar trajectories as the \emph{similar users} by two techniques:

i) $\epsilon$-thresholding: setting the threshold $\epsilon$ to filter similar users based on their trajectories' similarity. To be specific, if the trajectory similarity of $u_i$ and $u_j$ is greater than a threshold $\epsilon$, that is $SIM_t(u_i, u_j)>\epsilon$, this pair of users will be selected out as the \emph{similar users}, that is $u_i \backsim u_j$.
  
ii) clustering: grouping similar users together via {\em clustering} techniques. We use k-means clustering to cluster users based on their SSIM and EOTs features. We apply the Elbow and Silhouette method \cite{lleti2004selecting} to determine the number of clusters (k values). The resulting clusters present a group of highly similar users together. 

\subsubsection{Similarity of Outcome}
\label{section:sou}

To understand whether users with similar trajectories receive similar outcomes from the models, we first need to define what it means to receive a \emph{similar outcome} mathematically. 
As the objective of the PUT models is to optimize privacy gain and minimize utility loss, we consider privacy gain and utility gain as positive outcomes. After selecting out the similar users, we then measure the difference of them in privacy gain outcome, $\Delta D_{pri} = 1- D_{pri}(u_i)/D_{pri}(u_j)$, and utility gain outcome, $\Delta D_{uti}= 1-D_{uti}(u_i)/D_{uti}(u_j)$. 
Both $\Delta D_{pri}$ and $\Delta D_{uti}$ contribute to the evaluation of $SIM_o$. When with the \emph{clustering} approach, the average pairwise differences of $\Delta D_{pri}$ and $\Delta D_{uti}$ for all the members of each cluster are assessed.

Regardless of the grouping technique in similar users, we argue that $\Delta D_{pri}$ or $\Delta D_{uti}$ satisfies fairness if it is within $1-\epsilon$, otherwise, the PUT model is considered to be {\em violating individual fairness} for user pair $u_i$ and $u_j$. 
The threshold of different combinations of SSIM and EOTs are utilized to distinguish similar users and map all users into a list of \textit{pairs} with trajectory similarity and performance discrepancy.
To measure the fairness of systems as a whole for each model and outcome, we report the percentage of user pairs for whom fairness was violated (i.e., \textit{violation}\% or \textit{V}\%). As we will show, in our experiments, we set $\epsilon = 0.8$ to correspond to users with at least 80\% similarity of trajectory which imposes the model's outcome to be within 20\% difference between the similar users.  
The choice of $\epsilon=0.8$ is based on the various literature in fairness and literature~\cite{feldman2015certifying,barocas2017fairness}. We discuss the impact of this threshold on policy making in the Discussion section of this article.

\subsection{Group Fairness}

Different to individual fairness lies heavy on the similarity definition, group fairness has been vastly discussed and shares a systematic analysis approach in broader Fair-ML study. In this work, we bridge the gap between the standard group fairness metrics and the specific privacy-preserving mechanism of spatial-temporal data. 

Group fairness as also referred to as Demographic Parity~\cite{friedler2021possibility} states that demographic groups should receive similar decisions, inspired by civil rights laws in different countries~\cite{barocas2016big}.
To be specific, group fairness argues that a disadvantaged group (in terms of the sensitive attributes) should receive similar treatment to the advantaged group, that is:
\begin{equation}
\begin{aligned}
    P(Y'= 1|S = 0, Y = 1) = P(Y' = 1|S = 1, Y = 1)
\end{aligned}
\label{equ:group fairness}
\end{equation}

It is worth nothing that PUT spatial-temporal models are by definition group unaware that is $S$ indicating a sensitive attribute (e.g., race, or gender) is not an explicit feature into these models. However   specific demographic groups of users may exhibit certain properties in their mobility behaviour (e.g., students) that could still impact the  outcome of the PUT  models. 
For instance, age and employment status can highly influence peoples' day-to-day trajectory. A user whose trajectory data is limited to his home and office location could be highly predictable by the PUT model, however, also highly re-identifiable (with low privacy gain). This means the notion of group fairness in the context of this study is highly dependent on the examined \emph{dataset}. We elaborate more on this discussion in Section~\ref{sec:discussion}.

In order to quantify the group fairness in a more statistical approach, \textit{group fairness score} (i.e., \textit{GFS}) for spatial-temporal data are calculated by disparate impact for disadvantaged groups:
\begin{equation}
\begin{aligned}
    GFS = \frac{P(Y'= 1|S = advantaged, Y = 1)}{P(Y' = 1|S = disadvantaged, Y = 1)}
\end{aligned}
\label{equ:gfs}
\end{equation}
\section{Experiment Setup}\label{sec:setup}

In this section, we describe the datasets we used to evaluate the fairness of PUT models and the steps we took to set up the PUT models for examination.

\subsection{Datasets}

In order to evaluate the fairness of the examined models, we use two datasets that the original papers used to assess the privacy level of their models. 

\subsubsection{MDC}
This deadset is recorded from 2009 to 2011, contains a large amount of continuous mobility data for 184 volunteers with smartphones running a data collection software, in the Lausanne/Geneva area. 
Each record of the \textit{gps-wlan} dataset represents a phone call or an observation of a WLAN access point collected during the campaign ~\cite{laurila2012mobile}.  In addition to the trajectory data, MDC includes individual user demographic information: categorical age groups, gender, and employment status. To the best of our knowledge, MDC is the only dataset that has published users' demographic information along with their trajectories.

\subsubsection{Geolife}
This dataset is collected by Microsoft Research Asia from 182 users in the four-and-a-half-year period from April 2007 to October 2011 and contains 17,621 trajectories~\cite{zheng_geolife_2010}. As the Geolife dataset does not include demographic attributes of individuals, we are unable to measure the group fairness for this dataset and our analysis suffices for the individual fairness dimension. 

As mentioned in Section~\ref{section:sim of Tra}, in Figure~\ref{fig:heatmaps}, with the granularity coarsens, the trajectories become blurry and thus more similar to each other. 
Figure~\ref{fig:ssimgran} confirms this observation by illustrating the SSIM- and EOTs-based similarity of all the users for varying spatial granularity for both datasets. As the spatial granularity coarsens, we observe an increase in the SSIM values, with users becoming more similar to each other. Furthermore, as different types of entropy are considering different features of the spatial-temporal data, Figure~\ref{fig:ssimgran} presents the expected similarity of users for various EOTs-based measures. In addition to the distribution of the entropy values presented in the Figure~\ref{fig:ssimgran} for each dataset, we observe that across both datasets, SSIM along with SE and AE correspond to the most relaxed measure of similarity, LE and HE correspond to stricter measures of similarity. The corresponding percentage of user pairs that meet each similarity criterion is described in Table~\ref{tab:IF}.


\subsection{Original Properties of the Trajectory}

Before describing the privacy and utility trade-off for mobility trajectories of the PUT models, we first give brief definitions of two popular inference tasks (i.e., \emph{user re-identification and mobility prediction}), which are also applied to assess the privacy gain and utility decline in the PUT models we discussed. These two popular inference tasks are named \textit{original tasks} in this paper, where the \textit{original} demonstrates the nature of the data before being processed by any privacy-aware model. These \textit{original} tasks are leveraged to assess the native data characteristics in terms of \textit{user re-identification (UR)} and \textit{mobility predictability (MP)}, respectively. See the Appendix for full definitions.

\section{Fairness Analysis}\label{sec:results}
In this section, we present our analysis in studying whether the PUT models can be considered fair. To do so, we analyze these models in terms of individual fairness and group fairness. The similarity $SIM_t$ applied in the individual fairness is defined by SSIM and EOTs, and group fairness is grouping users based on demographic attributes such as gender, age, and employment status.

\subsection{Individual Fairness}
The metrics of trajectories' similarity $SIM_t$ are crucial for quantifying individual fairness. As definitions in Section~\ref{section:IF}, the $SIM_t$ can be quantified by SSIM and EOTs. In this section, we discuss individual fairness with two different similarity quantification approaches. First, the $SIM_t$ discriminated based on $\epsilon$-thresholding metrics of SSIM and EOTs directly. Second, the k-means clustering approach, based on the characteristics of SSIM and EOTs aforementioned, is leveraged to classify similar users.

\subsubsection{Similarity Based on $\epsilon$-Thresholding}
\label{section:if_one}

\begin{table*}[t]\small
   \centering
   \begin{tabular}{ccc|cccccc}
    \hline
    & \multirow{3}{*}{\shortstack{Metrics}} & \multirow{3}{*}{\shortstack{\% of pairs}} & \multicolumn{2}{c|}{Original, V\% of (DIFF>0.2)} & \multicolumn{2}{c|}{Mo-PAE, V\% of (DIFF>0.2)} & \multicolumn{2}{c}{TrajGAN, V\% of (DIFF>0.2)} \\
    \cline{4-9}
       &  &  & \multicolumn{1}{c}{\multirow{2}{*}{\shortstack{Trajectory \\ Uniqueness}}} & \multicolumn{1}{c|}{\multirow{2}{*}{\shortstack{Mobility \\ Predictability}}}  & \multirow{2}{*}{\shortstack{Privacy \\ Gain}}  & \multicolumn{1}{c|}{\multirow{2}{*}{\shortstack{Utility \\ Decline}}}  & \multirow{2}{*}{\shortstack{Privacy \\ Gain}} & \multicolumn{1}{c}{\multirow{2}{*}{\shortstack{Utility \\ Decline}}} \\
       & & & & \multicolumn{1}{c|}{} & & \multicolumn{1}{c|}{} &  \\
       
       \cline{1-9}
       \multirow{8}{*}{\shortstack{MDC}} 
       & SE & 36.17\% & \textit{10.50\%} & \textit{11.11\%} & 87.69\% & 39.75\% & 41.65\% & 27.32\% \\
       & LE & 12.85\% & \textit{8.31\%} & \textit{7.90\%} & 88.81\% & 36.95\% & 41.32\% & {\bf 25.10\%} \\
       & HE & 14.11\% & \textit{12.89\%} & \textit{9.60\%} & 86.88\% & 41.30\% & {\bf 38.23\%} & 27.14\% \\       
       & AE & 33.05\% & \textit{12.64\%} & \textit{10.28\%} & 87.10\% & 35.95\% & \textbf{45.26\%} & \textbf{29.42\%} \\
       \cdashline{2-9}[1pt/3pt]
        & SSIM & 65.06\% & \textbf{\emph{14.57\%}} & \textbf{\emph{13.17\%}} & \textbf{88.98\%} & \textbf{42.02\%} & 39.50\% & 27.50\% \\
       \cdashline{2-9}[1pt/3pt]
       & EOTs & 1.73\% & \textit{\textbf{6.10\%}} & \textit{\textbf{1.22\%}} & 84.76\% & \textbf{30.49\%} & 44.51\% & 27.44\% \\
       & EOTs+SSIM & 1.64\% & \textit{6.45\%} & \textit{1.29\%} & {\bf 83.87\%} & 31.61\% & 43.23\% & 28.39\% \\
       \cline{1-9}
       \multirow{8}{*}{\shortstack{Geolife}} 
       & SE & 33.16\% & 57.91\% & 61.09\% & 94.14\% & 71.84\% & 67.85\% & 58.58\% \\
       & LE & 9.11\% & {\bf 57.41\%} & 61.20\% & {\bf 94.32\%} & 71.50\% & 65.09\% & 56.05\% \\
       & HE & 7.29\% & 61.37\% & \textbf{63.47\%} & 94.09\% & 70.43\% & 69.78\% & 58.87\% \\
       & AE & 27.03\% &  57.44\% & 59.36\% & 93.23\% & 72.96\% & 71.96\% & 58.54\% \\
       \cdashline{2-9}[1pt/3pt]
       & SSIM & 63.52\% & 59.88\% & 61.49\% & 94.13\% & \textbf{74.77\%} & {\bf 63.53\%} & \textbf{53.05\%} \\
       \cdashline{2-9}[1pt/3pt]
       & EOTs & 0.62\% & 61.54\% & 58.46\% & {\bf 89.23\%} & 66.15\% & \textbf{78.46\%} & \textbf{72.31\%} \\
       & EOTs+SSIM & 0.61\% & \textbf{62.50\%} & {\bf 57.81\%} & 89.06\% & {\bf 65.63\%} & 78.13\% & 71.88\% \\
        \hline
    \end{tabular}
    
    \caption{Individual fairness among diverse models and datasets with SSIM and EOTs.
    \textit{\% of pairs} represents the ratio of the pairs that meet the thresholding requirements.
    The maximum/minimum instances of each column are highlighted in \textbf{bold font}.
    }
    \label{tab:IF}
\end{table*}

Table~\ref{tab:IF} presents the individual fairness of different models by the $\epsilon$-thresholding metrics based on SSIM and EOTs. The threshold $\epsilon$ of different combinations of SSIM and EOTs are utilized to distinguish similar users ($u_i \backsim u_j$) and map all users into a list of \textit{pairs} with trajectory similarity and performance discrepancy. Based on  fairness thresholding criteria defined in Section~\ref{section:sou}, \textit{similar users} (i.e., \textit{user pairs}) imply at least 80\% pairwise similarity of their trajectories. 
\textit{"\% of pairs"} in the table represents the percentage of the user pairs that meet the corresponding metric threshold requirements. For instance, with the MDC dataset, 36.17\% of \textit{user pairs} have a more than 80\% similarity when under the $SE$ metric. That is, under the $SE$ metric, 36.17\% user pairs are qualified for further analysis of outcome similarity.

The \textit{user pair} is defined to achieve individual fairness when the outcome difference ($\Delta D_{pri}$ or $\Delta D_{uti}$) between $u_i$ and $u_j$ is within 20\%. 
Table~\ref{tab:IF} shows the percentage of \textit{user pairs} that commit fairness violation (i.e., V\% = \% of ($\Delta D$>0.2)). 
For instance, in Table~\ref{tab:IF}, with the MDC dataset under the $SE$ metric, there are only 10.50\% and 11.11\% of the \textit{qualified user pairs} violate the fairness criteria in two original tasks, which implies that the individual fairness is achieved, as both V\% are within 20\%.
Different from the original tasks, two PUT models have V\% that are all higher than 20\%, hence, they violate individual fairness.
The higher V\% indicates that the model causes more disparities in performance.
The values in the \textit{italic format} present the cases where the outcome to meet individual fairness (i.e., $V\% \leq 20\%$) in the Table~\ref{tab:IF}.

Overall, individual fairness is {\bf not achieved} in the two selected PUT models, especially for the unfairness of the privacy gain, which is generally higher than the utility decline. When comparing two different privacy models in a row, TrajGAN achieves less fairness violation rate than Mo-PAE in both privacy gain and utility decline outcomes. For instance, in the MDC dataset, when 45.26\% and 29.42\% of user pairs commit fairness violations in privacy gain and utility decline, respectively, the Mo-PAE reports twice as many fairness violations for both outcomes.
While both the Geolife and MDC data exhibit individual unfairness, the Geolife is worse in both the PUT models and the accuracy of the \textit{original tasks}. 
In both original tasks, Geolife's unfairness rate is as high as 60\%, and this inequity is exacerbated when with PUT models.
In contrast to Geolife, the performance of the MDC in the original tasks conforms to the definition of individual fairness, that is, the performance difference of task accuracy in MDC is within 20\% in both user re-identification tasks and mobility prediction tasks.

\textbf{Impact of Spatial Granularity on Similarity:} 
After the overall comparison of threshold metrics, we discuss the model discrepancy when trajectory similarity is based on the SSIM index under varying granularity. As a crucial metric in distinguishing the trajectory similarity, the SSIM index could be affected by different parameters, which will result in subtle performance disparities in the quantification of individual fairness. The spatial granularity of trajectory is the most important one among these parameters. 
These disparities could be intuitively observed in the heatmaps (Figure~\ref{fig:heatmaps}).
In contrast to the SSIM, the spatial granularity has less impact on different types of entropy, hence, they are not discussed here.

\begin{figure*}
     \centering

     \begin{subfigure}[b]{0.95\textwidth}
         \centering
         \includegraphics[width=\textwidth]{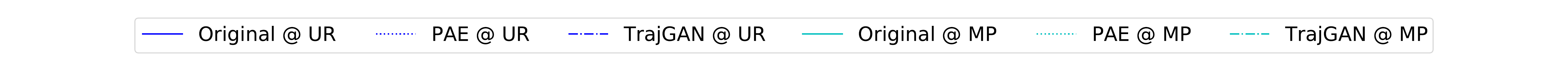}
     \end{subfigure}
     \hfill
     \begin{subfigure}[b]{0.20\textwidth}
         \centering
         \includegraphics[width=\textwidth]{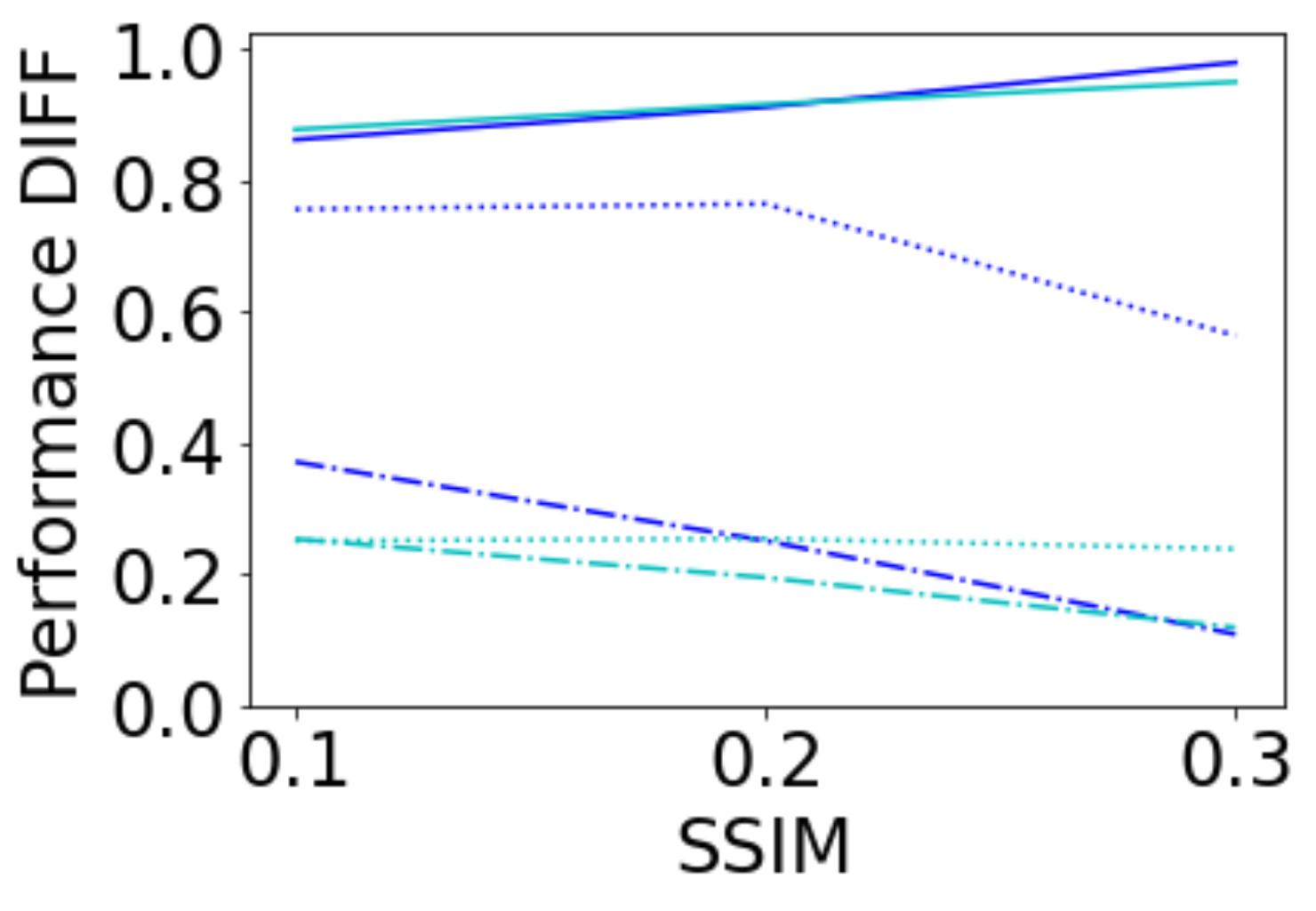}
         \caption{100m, MDC}
         \label{fig:mdc_100_xssim}
     \end{subfigure}
     \hfill
     \begin{subfigure}[b]{0.20\textwidth}
         \centering
         \includegraphics[width=\textwidth]{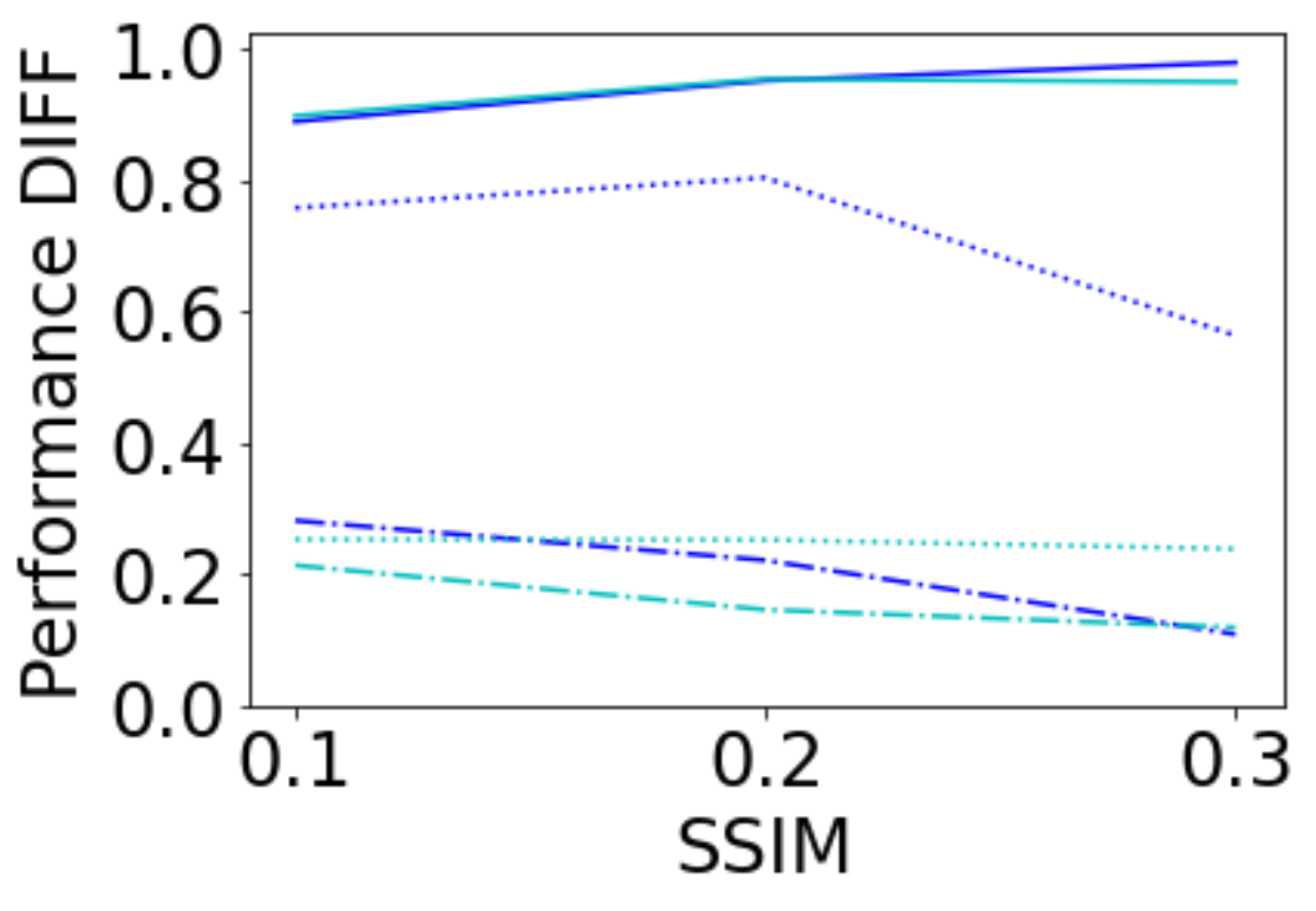}
         \caption{300m, MDC}
         \label{fig:mdc_300_xssim}
    \end{subfigure}
    \hfill
    \begin{subfigure}[b]{0.20\textwidth}
         \centering
         \includegraphics[width=\textwidth]{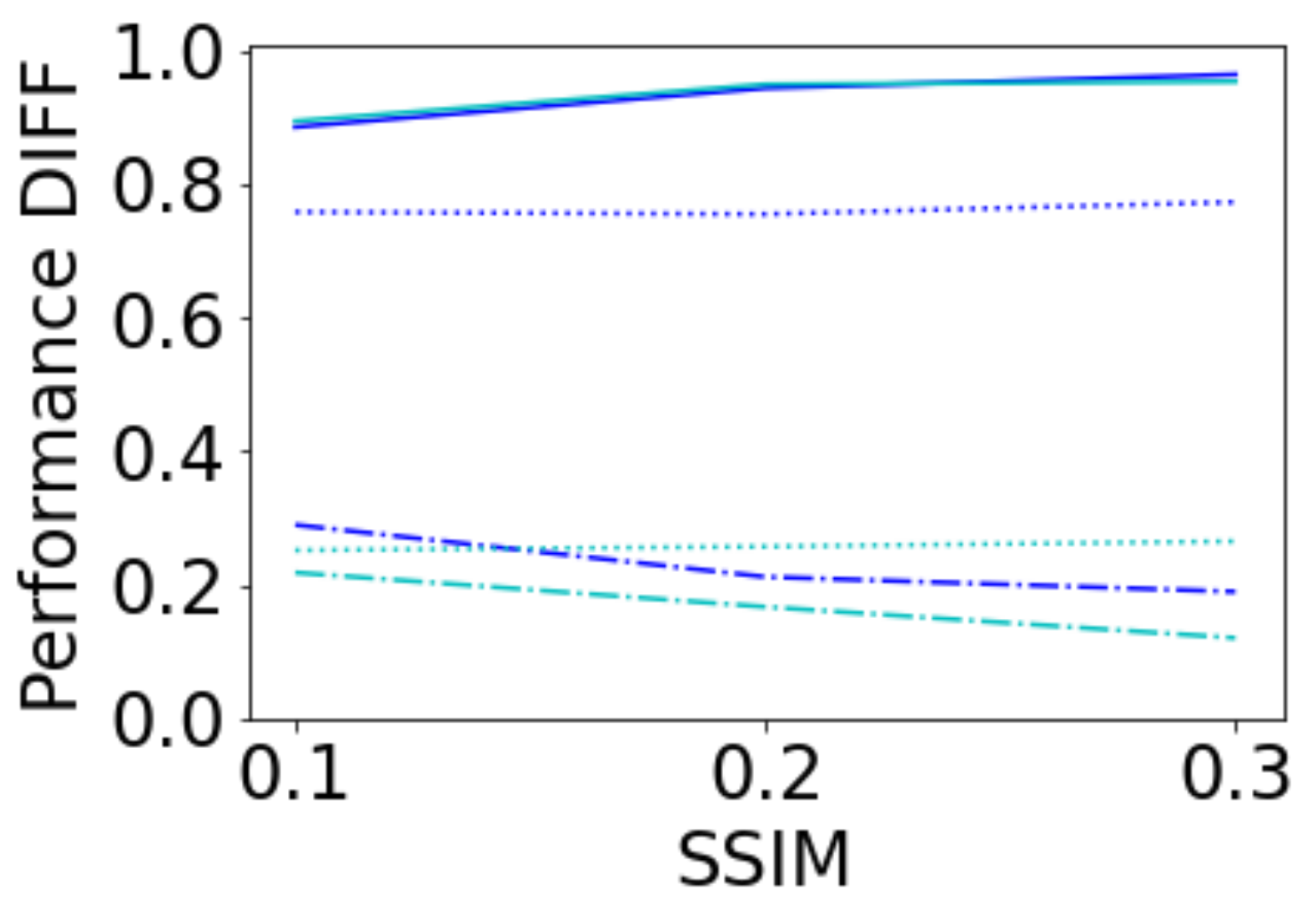}
         \caption{500m, MDC}
         \label{fig:mdc_500_xssim}
     \end{subfigure}
     \hfill
    \begin{subfigure}[b]{0.20\textwidth}
         \centering
         \includegraphics[width=\textwidth]{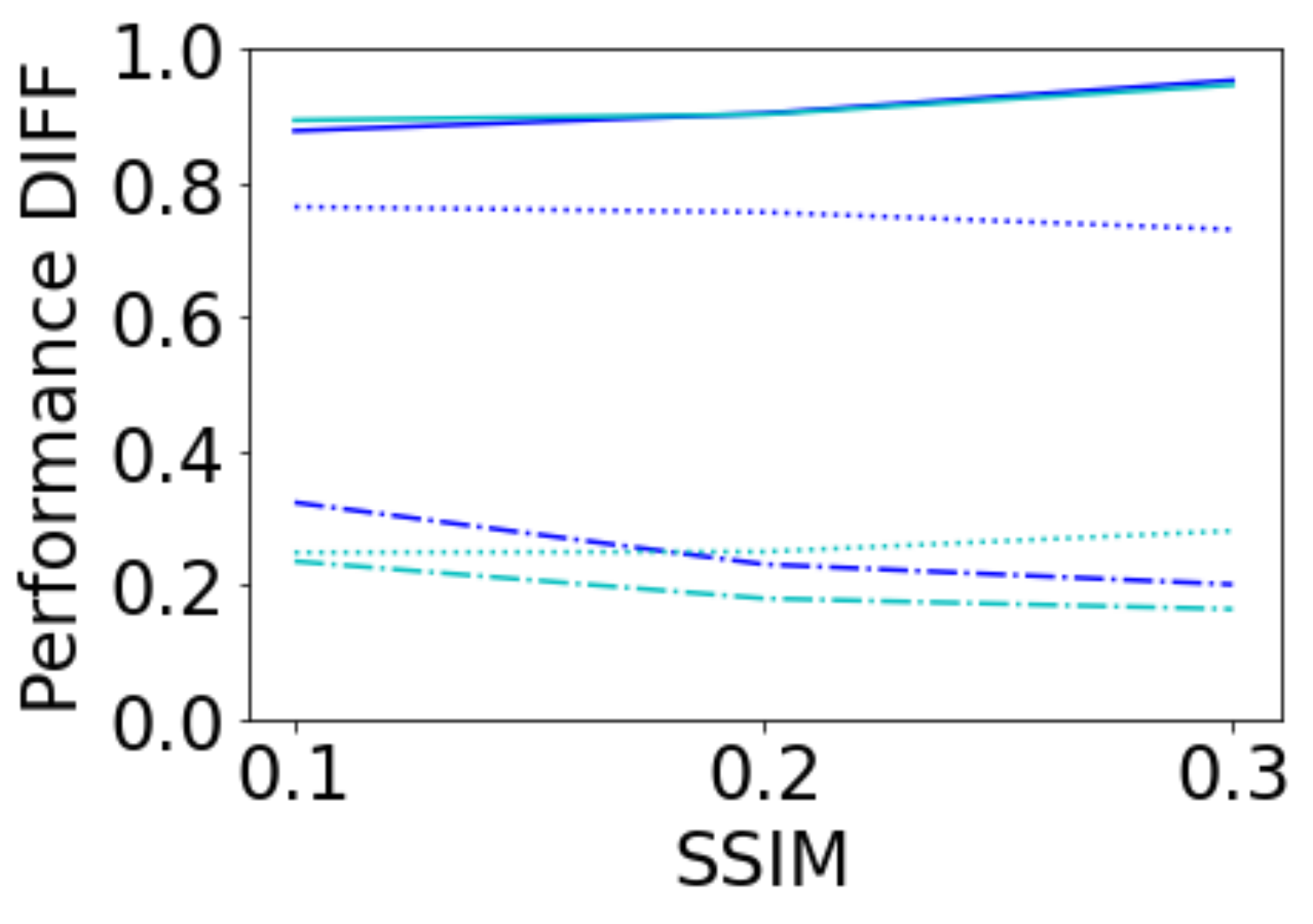}
         \caption{900m, MDC}
         \label{fig:mdc_900_xssim}
     \end{subfigure}
     \begin{subfigure}[b]{0.20\textwidth}
         \centering
         \includegraphics[width=\textwidth]{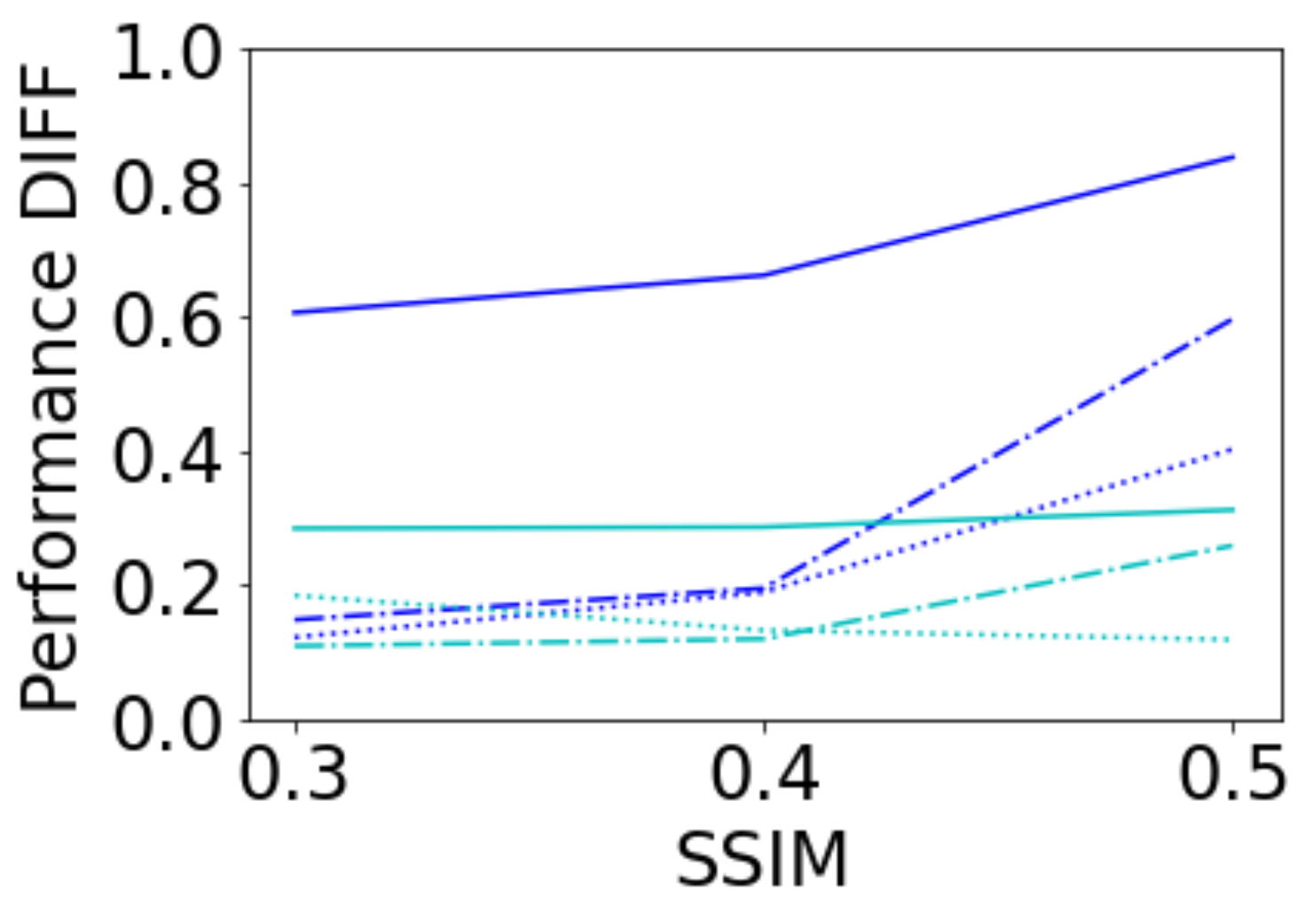}
         \caption{100m, Geolife}
         \label{fig:geo_100_xssim}
     \end{subfigure}
     \hfill
     \begin{subfigure}[b]{0.20\textwidth}
         \centering
         \includegraphics[width=\textwidth]{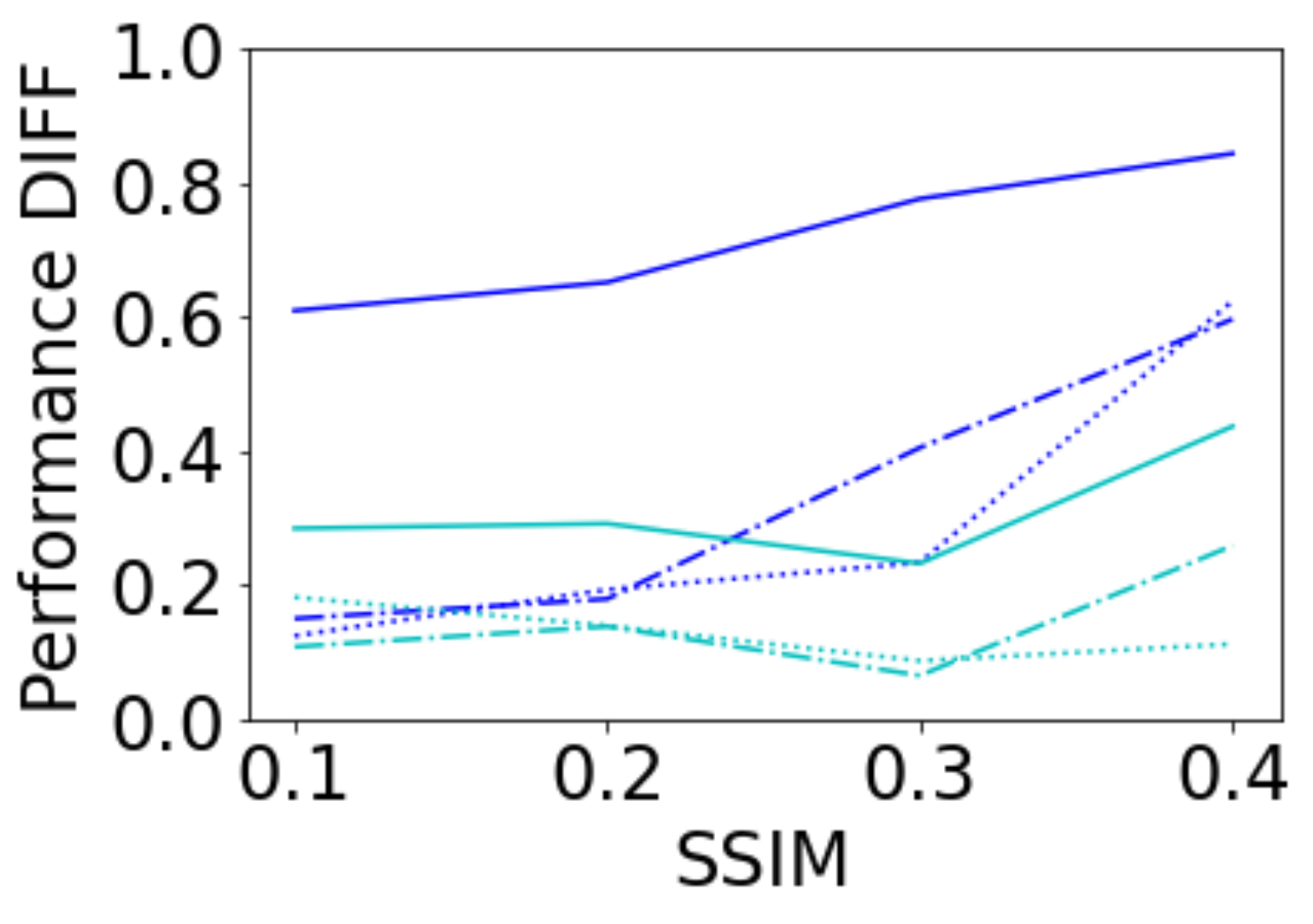}
         \caption{300m, Geolife}
         \label{fig:geo_300_xssim}
    \end{subfigure}
    \hfill
    \begin{subfigure}[b]{0.20\textwidth}
         \centering
         \includegraphics[width=\textwidth]{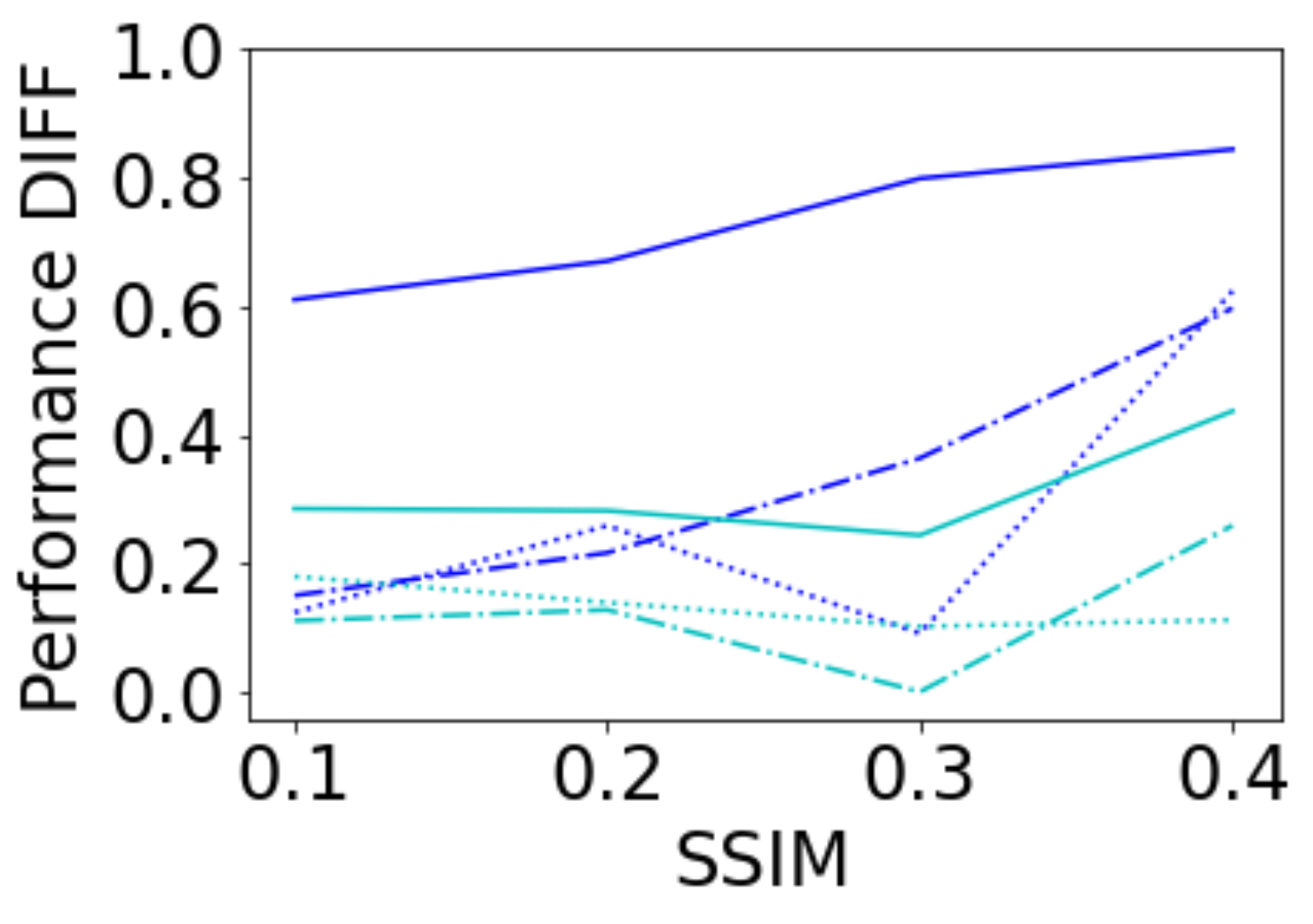}
         \caption{500m, Geolife}
         \label{fig:geo_500_xssim}
     \end{subfigure}
     \hfill
    \begin{subfigure}[b]{0.20\textwidth}
         \centering
         \includegraphics[width=\textwidth]{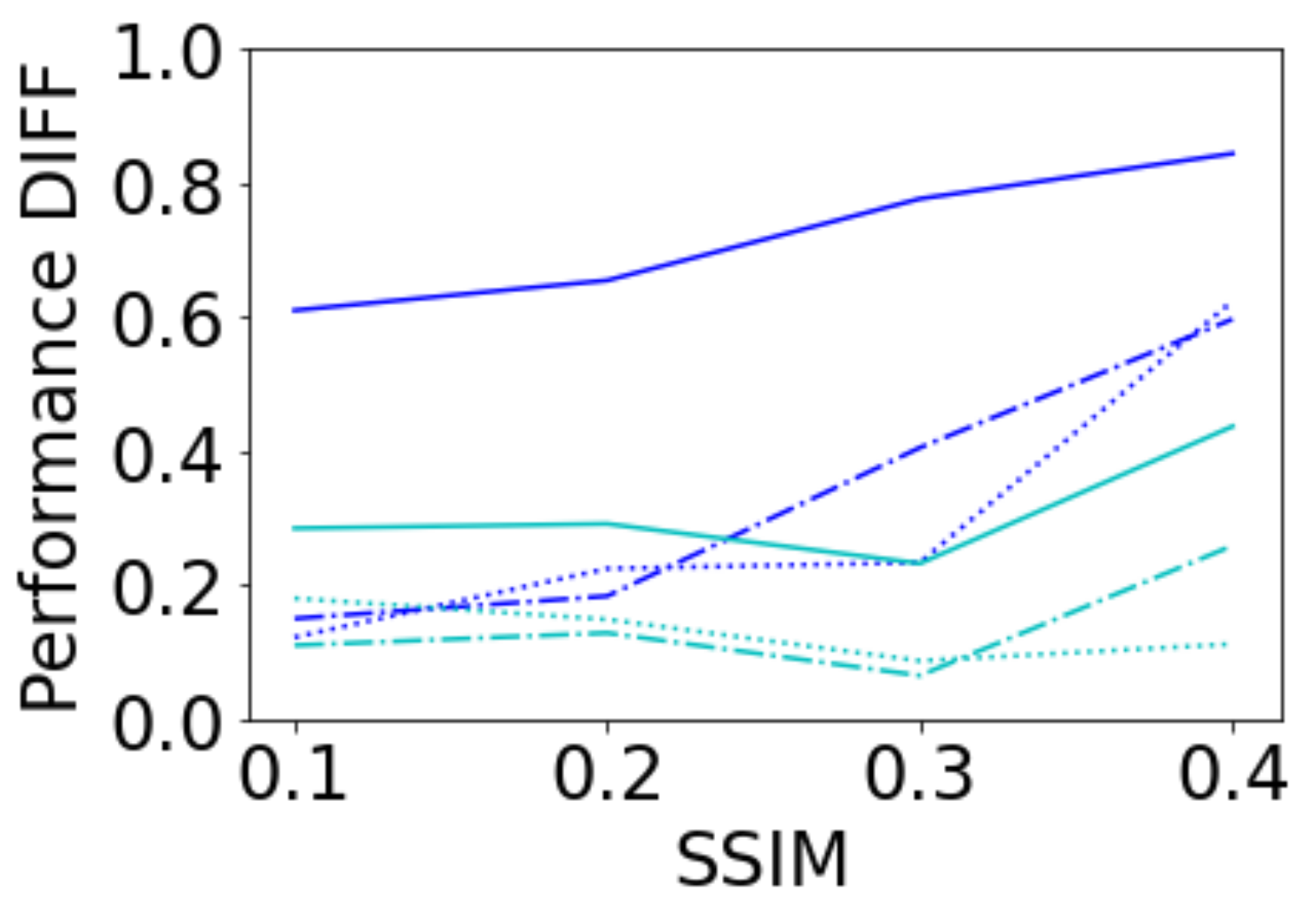}
         \caption{900m, Geolife}
         \label{fig:geo_900_xssim}
     \end{subfigure}
     \hfill
    \caption{The model performance discrepancy when trajectory similarity is based on the SSIM in different granularities. Figure (a) to Figure (d) are the results of MDC dataset, Figure (e) to Figure (h) are of Geolife.
    The performance discrepancy (i.e., Performance DIFF) of each model in different granularities compares in each sub-figure.}
    \label{fig:xssim_gran}
\end{figure*}

The Figure~\ref{fig:xssim_gran} then shows the impact of varying spatial granularity on the model discrepancy. The model that achieves individual fairness should perform less discrepancy with higher SSIM. The accuracy of original tasks and two PUT models are compared in granularity at 100 meters, 300 meters, 500 meters, and 900 meters.
In conclusion, different models have diverse sensitivities of varying granularities. 
Both original tasks (UR and MP) in the two datasets have an increasing difference with a higher SSIM index, which means they violate individual fairness. 
For the Mo-PAE, individual fairness is met on MDC data but not on Geolife. The Mo-PAE is also the most sensitive model for varying granularities. For instance, when granularity changes from 100-meter (Figure~\ref{fig:mdc_100_xssim}) to 900-meter (Figure~\ref{fig:mdc_900_xssim}), Mo-PAE has the most obvious change in its line trend on the UR (i.e., privacy gain), and the decreasing trend at 100-meter granularity is lost at 900-meter. 
Overall, the selection of SSIM granularity has a significant impact on the judgement of the individual fairness of a model. However, these impacts become subtle when the SSIM is applied to the trajectory similarity distinction, as the user pairs table reduced the granularity impact to some extent. For the remaining of the analysis, the granularity of the SSIM is chosen as 100-meter.

\subsubsection{Similarity Based on K-means Clustering}
\begin{table*}[t]\small
   \centering
   \begin{tabular}{ccc|cccccc}
    \hline
    & \multirow{3}{*}{\shortstack{Metrics}} & \multirow{3}{*}{\shortstack{Cluster \\ Size}} & \multicolumn{2}{c|}{Original, V\% of (DIFF>0.2)} & \multicolumn{2}{c|}{Mo-PAE, V\% of (DIFF>0.2)} & \multicolumn{2}{c}{TrajGAN, V\% of (DIFF>0.2)} \\
    \cline{4-9}
       &  &  & \multicolumn{1}{c}{\multirow{2}{*}{\shortstack{Trajectory \\ Uniqueness}}} & \multicolumn{1}{c|}{\multirow{2}{*}{\shortstack{Mobility \\ Predictability}}}  & \multirow{2}{*}{\shortstack{Privacy \\ Gain}}  & \multicolumn{1}{c|}{\multirow{2}{*}{\shortstack{Utility \\ Decline}}}  & \multirow{2}{*}{\shortstack{Privacy \\ Gain}} & \multicolumn{1}{c}{\multirow{2}{*}{\shortstack{Utility \\ Decline}}} \\
       & & & & \multicolumn{1}{c|}{} & & \multicolumn{1}{c|}{} &  \\
       \cline{1-9}
      \multirow{5}{*}{\shortstack{MDC}} 
       & Cluster 1 & 26 & \textit{14.77\%}          & \textit{15.38\%}           & {\bf 83.69\%} & {\bf 48.92\%}            & 50.77\%       & 33.85\%                  \\
       & Cluster 2 & 5  & \textit{\textbf{0.00\%}}  & \textit{\textbf{0.00\%}}   & {\bf 90.00\%} & \textit{\textbf{0.00\%}} & 40.00\%       & \textit{\textbf{0.00\%}} \\
       & Cluster 3 & 43 & \textit{\textbf{17.39\%}} & \textit{\textbf{17.17\%}}  & 88.15\%       & 48.84\%                  & {\bf 55.92\%} & {\bf 43.63\%}            \\
       & Cluster 4 & 24 & \textit{\textbf{0.00\%}}  & \textit{\textbf{0.00\%}}   & 86.23\%       & \textit{16.67\%}         & {\bf 35.87\%} & \textit{17.75\% }        \\
       & Clusters Average & - & \textit{12.99\%}    & \textit{12.18\%}           & 88.86\%       & 39.36\%                  & 51.26\%       & 34.49\%                  \\
       \cline{1-9}
       \cline{1-9}
       \multirow{6}{*}{\shortstack{Geolife}}
       & Cluster 1 & 21 & \textbf{55.71\%}          & \textit{\textbf{13.81\%}} & {\bf 60.00\%} & \textit{1.43\%}          & \textit{18.57\%}          & \textit{\textbf{0.00\%}} \\
       & Cluster 2 & 17 & 46.32\%                   & \textit{8.09\%}           & 49.26\%       & \textit{13.97\%}         & \textit{16.91\%}          & \textit{\textbf{0.00\%}} \\
       & Cluster 3 & 9  & \textit{\textbf{13.89\%}} & \textit{11.11\%}          & 38.89\%       & {\bf 16.67\%}            & \textit{\textbf{13.89\%}} & \textit{\textbf{11.11\%}} \\
       & Cluster 4 & 10 & 31.11\%                   & \textit{\textbf{0.00\%}}  & 40.00\%       & \textit{\textbf{0.00\%}} & {\bf 44.44\%}.            & \textit{4.44\%} \\
       & Cluster 5 & 36 & 44.92\%                   & \textit{8.57\%}           & {\bf 29.84\%} & \textit{5.56\%}          & 23.02\%                   & \textit{0.16\%} \\
       & Clusters Average & - & 43.91\%             & \textit{9.15\%}           & 47.11\%       & \textit{7.55\%}          & 26.53\%                   & \textit{2.02\%} \\
       \hline
    \end{tabular}
    \caption{K-means-clustering-based individual fairness among diverse models and datasets. The numbers present the percentage of users for whom individual fairness was violated based on their difference in the outcome being greater than 0.2. The fair instances are highlighted in \textit{italic font}. The maximum/minimum instances of each column are highlighted in \textbf{bold font}.}
    \label{tab:clusters}
\end{table*}


Alternative to the results presented based on the similarity thresholding, Table~\ref{tab:clusters} demonstrates the results of individual fairness based on the clustering technique described in Section~\ref{section:sou}. Applying the Elbow and Silhouette methods, we decide the number of clusters (k) to be  4 and 5 for  MDC and Geolife, respectively.
For each cluster, the table reports the percentage of users whose individual fairness was violated for a given outcome and under various models. More precisely, the results presented here indicate that the original model that objectifies a single task (prediction or privacy) is able to meet the individual fairness criteria for the MDC dataset.  We can observe that in the case of the Mo-PAE model, the privacy gain  exhibits high variations across users in the same clusters.  Even in the cases where the model satisfies individual fairness by performing similarly in terms of utility decline (clusters 2 and 4 of MDC and all clusters in Geolife ), the privacy gains of those users are very different from each other.

\begin{figure*}
     \centering
     \hfill
     \begin{subfigure}[b]{0.30\textwidth}
         \centering
         \includegraphics[width=\textwidth]{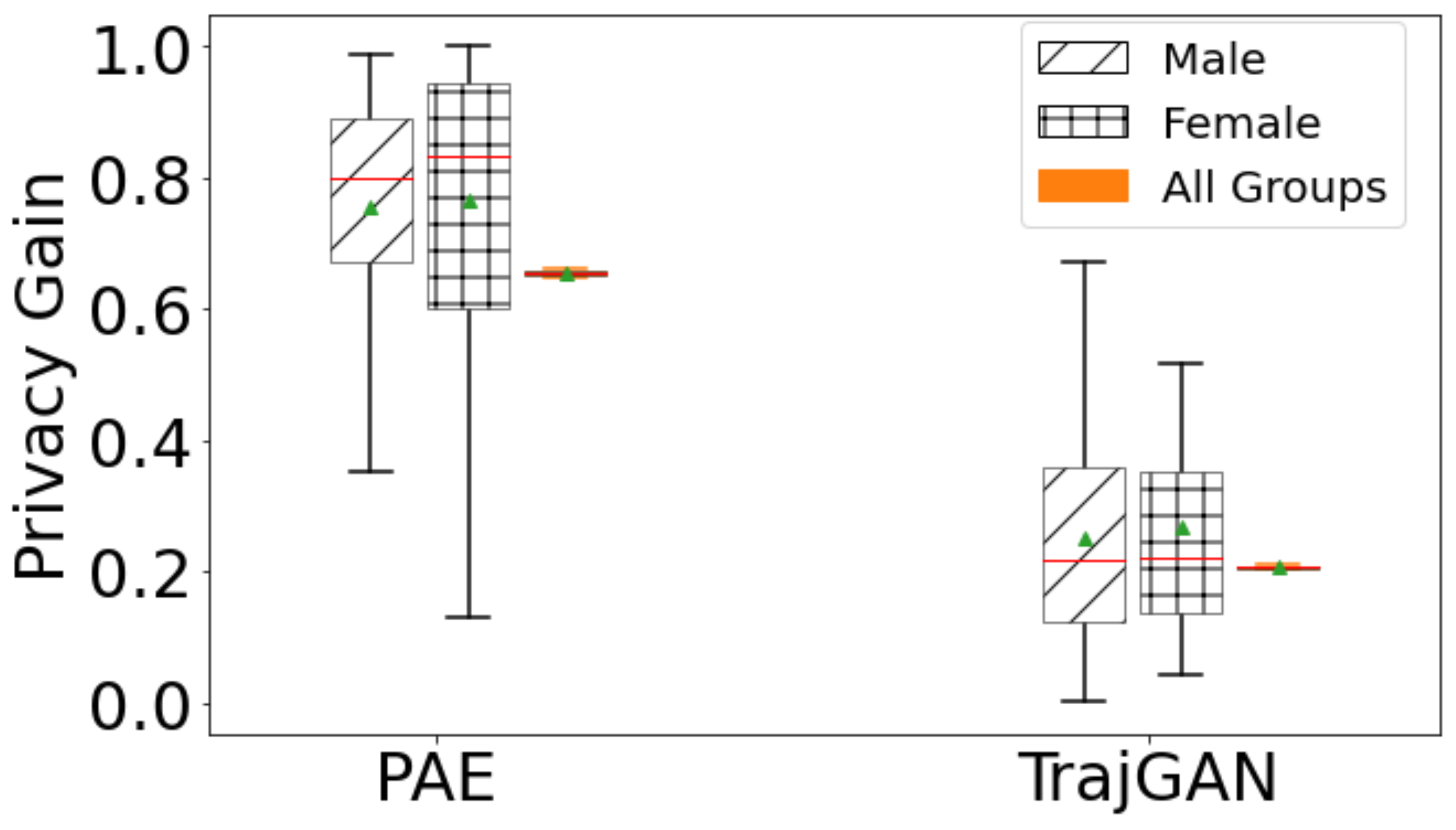}
         \caption{Gender}
         \label{fig:gender_p}
     \end{subfigure}
     \hfill
     \begin{subfigure}[b]{0.30\textwidth}
         \centering
         \includegraphics[width=\textwidth]{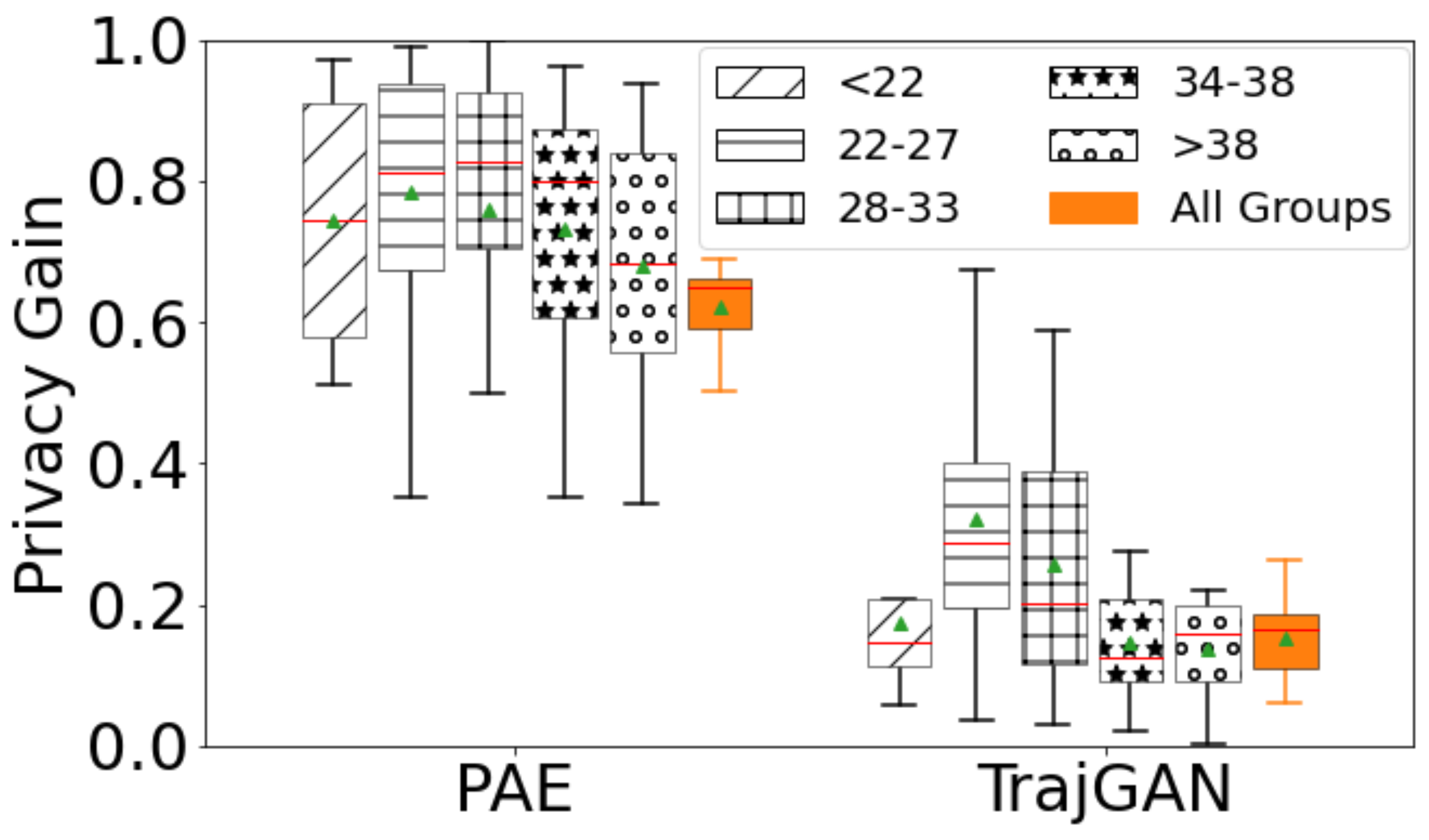}
         \caption{Age}
         \label{fig:age_p}
    \end{subfigure}
    \hfill
    \begin{subfigure}[b]{0.30\textwidth}
         \centering
         \includegraphics[width=\textwidth]{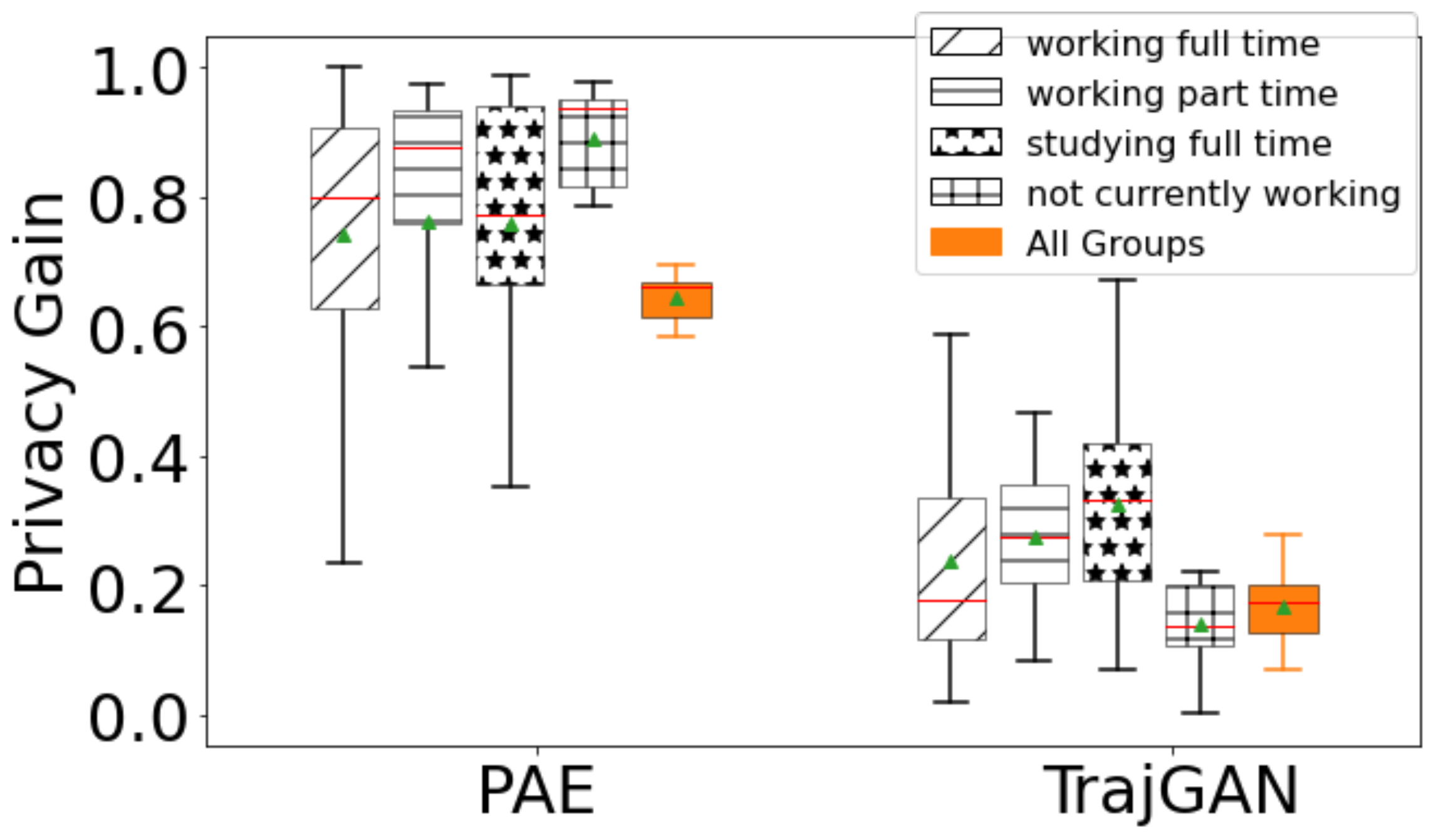}
         \caption{Working Status}
         \label{fig:work_p}
     \end{subfigure}
     \hfill
    \caption{The privacy protection outcome of PUT models across different demographic groups for the MDC dataset. 
    }
    \label{fig:group_privacy}
\end{figure*}

\begin{figure*}
     \centering
     \hfill
     \begin{subfigure}[b]{0.30\textwidth}
         \centering
         \includegraphics[width=\textwidth]{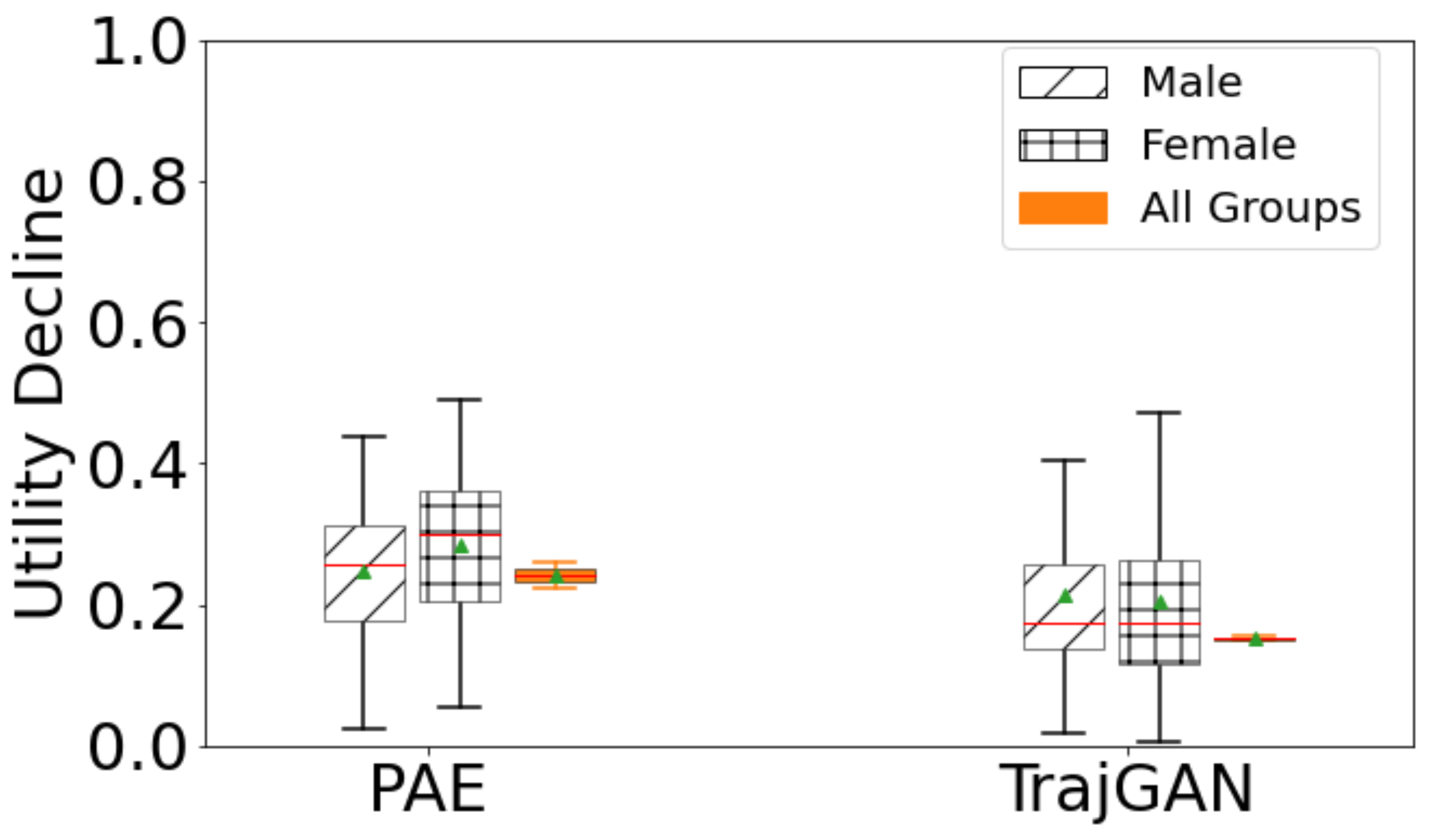}
         \caption{Gender}
         \label{fig:gender_u}
     \end{subfigure}
     \hfill
     \begin{subfigure}[b]{0.30\textwidth}
         \centering
         \includegraphics[width=\textwidth]{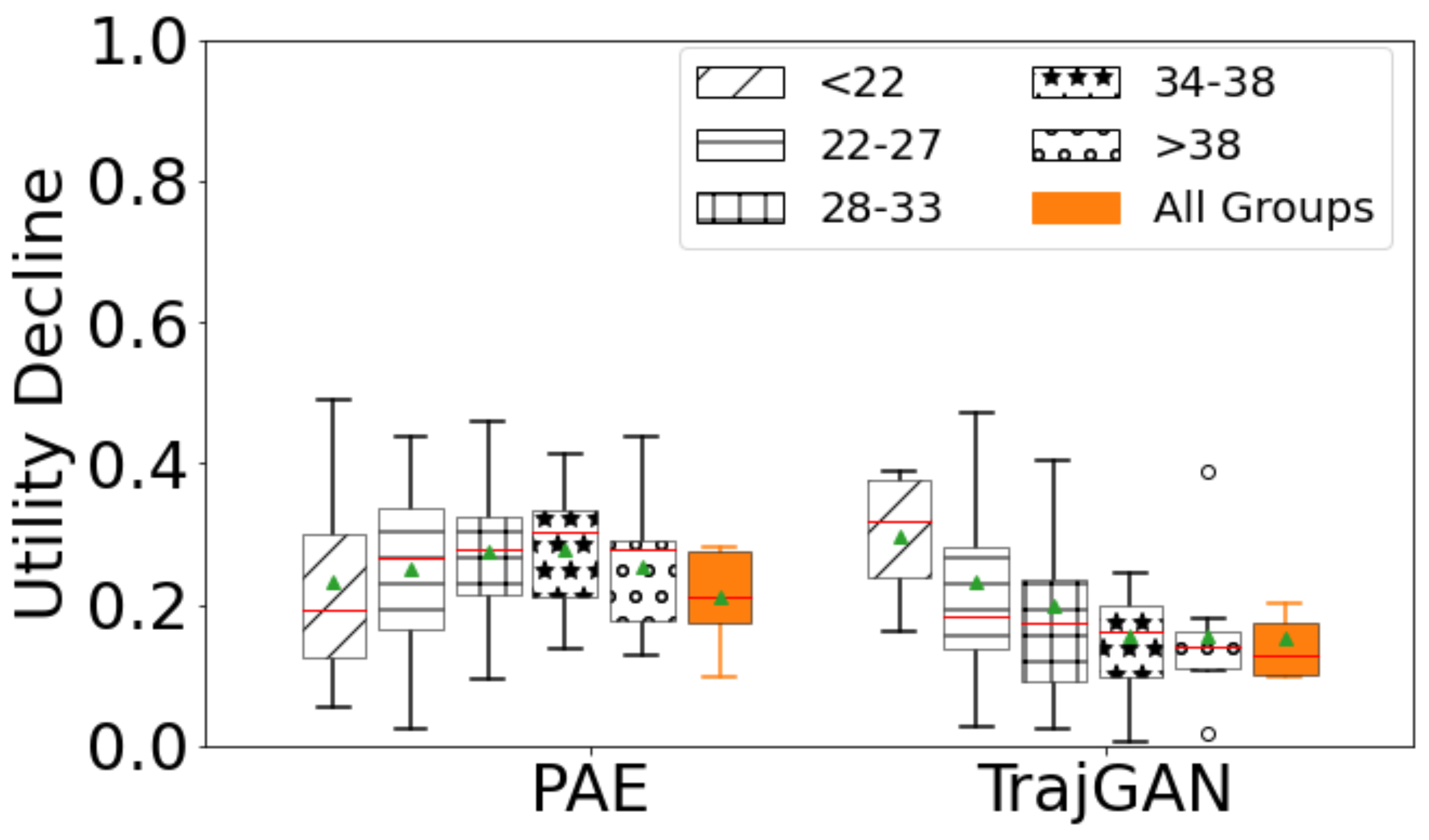}
         \caption{Age}
         \label{fig:age_u}
    \end{subfigure}
    \hfill
    \begin{subfigure}[b]{0.30\textwidth}
         \centering
         \includegraphics[width=\textwidth]{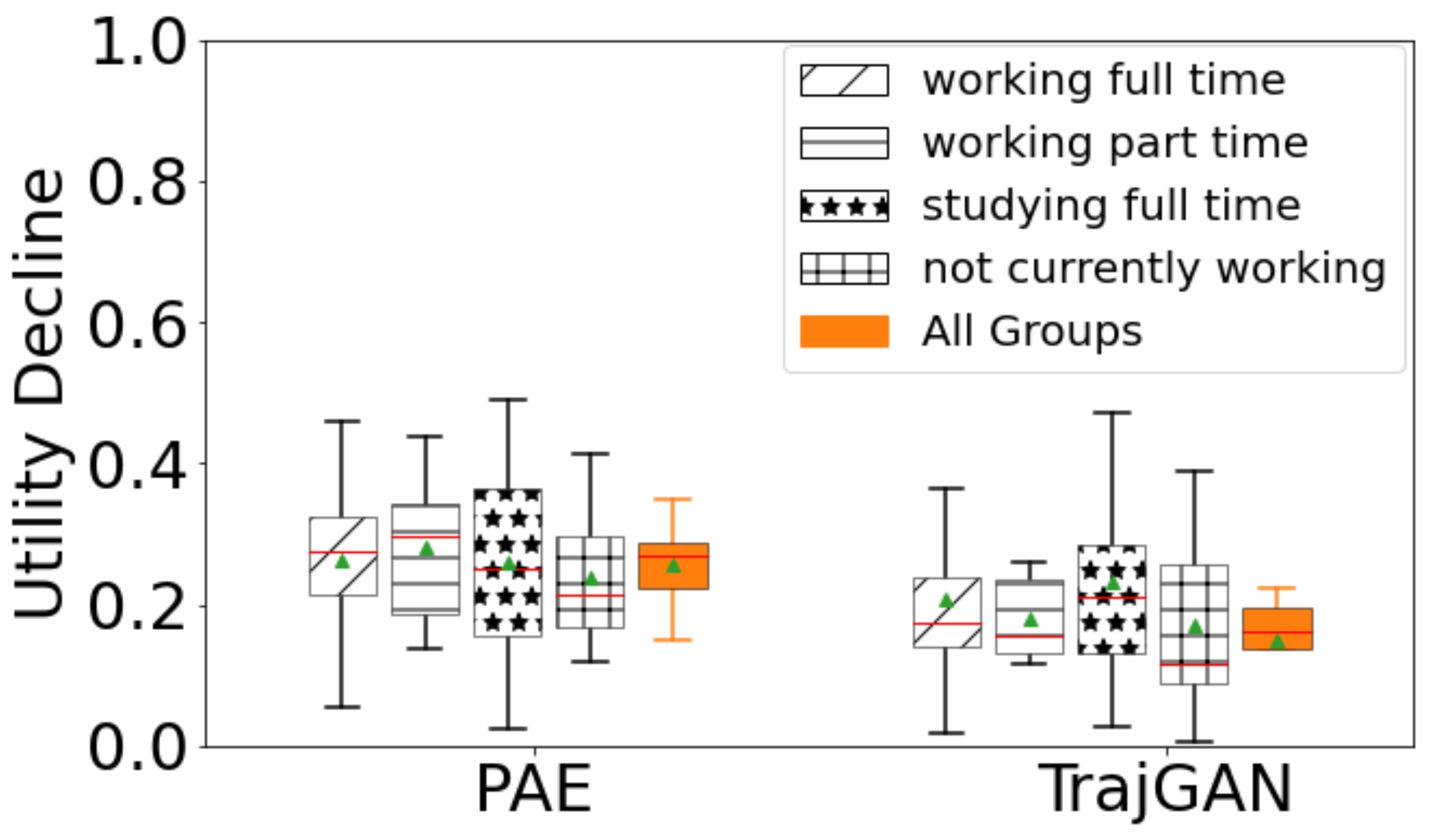}
         \caption{Working Status}
         \label{fig:work_u}
     \end{subfigure}
     \hfill
    \caption{The prediction accuracy outcome of PUT models across different demographic groups for the MDC dataset.}
    \label{fig:group_utility}
\end{figure*}

\subsection{Group Fairness}

Group fairness states that  groups across different sensitive attributes should receive similar outcomes.
To be specific, group fairness argues that a disadvantaged group should receive similar treatment to the advantaged group.
Figure~\ref{fig:group_privacy} presents the discrepancy of the \textit{privacy gain} from two PUT models for different demographic groups, and Figure~\ref{fig:group_utility} presents the {\em utility decline}. 
We observe that both Mo-PAE and TrajGAN perform equally for different gender attributes, as shown in Figure~\ref{fig:gender_p}, where the orange boxes (labelled as \textit{All Groups}) on both are very small. That is while the privacy gain varies across individuals within the same gender, the model achieves group fairness when grouping individuals by gender.  
The same observations could be made for the age and employment status, where we see that there exist bigger differences across the classes than the gender, but they still achieve group fairness as $\Delta D <20\%$. 
Similarly, in Figure~\ref{fig:group_utility}, we can observe that both models equally meet the group fairness criteria on the utility decline. 

In order to quantify the group fairness of the \textit{disadvantaged groups} in a more statistical approach, the results of the \textit{group fairness score} (\textit{GFS}) are shown in Table~\ref{tab:GF}.
For instance, for different age groups, the subgroup with ages between 22 and 27 (i.e., "22 - 27") is regarded as the \textit{advantaged} group, as it has the dominant user number for all age groups. The other age groups' GFSs are calculated based on the disparate impact between them and the \textit{advantaged} group. Then, compare all GFSs against the fairness threshold of 0.8, which is defined in Section~\ref{section:sou}, that is, \textit{GFS} $\geq 80\%$ indicates fairly treating the disadvantaged group and \textit{GFS} $< 80\%$ indicates the unfairly treating. For example, the result of "28-33" group (i.e., \textit{GFS} $= 98.65\%$) then indicated that the model satisfy the group fairness as 98.65\% > 80\%.

In conclusion, except for two subgroups with age attributes (i.e., "<21" and ">39") violating the four-fifths rule, the other subgroups satisfy the group fairness.  Finally, it is worth noting that the results presented here are highly dependent on the studied dataset, as we discuss in the next section. 

\begin{table*}[t]\small
   \centering
   \begin{tabular}{ccccccccc}
    \hline
   & & \multirow{2}{*}{\shortstack{users \#}} & \multicolumn{2}{|c|}{Original, GFS} & \multicolumn{2}{c|}{Mo-PAE, GFS} & \multicolumn{2}{c}{TrajGAN, GFS} \\
    \cline{4-9}
      &  &  &  \multicolumn{1}{|c}{Uniqueness} & \multicolumn{1}{c|}{Predictability}  & PrivacyGain  & \multicolumn{1}{c|}{UtilityDecline}  & PrivacyGain & \multicolumn{1}{c}{UtilityDecline}\\
    \cline{1-9}
    \multirow{2}{*}{\shortstack{Gender}} & Male & 56 & - & - & - & - & - & - \\
       & Female & 33 & 98.07\% & 96.35\% & 98.04\% & 90.13\% & 96.57\% & 95.00\% \\
    \hline
    \multirow{5}{*}{\shortstack{Age}} & <21 & 5 & 94.48\% & 99.10\% & \textbf{46.09\%} & 85.86\% & 84.73\% & 87.38\% \\
      & 22-27 & 38 & - & - & - & - & - & - \\
      & 28-33 & 29 & 98.65\% & 94.49\% & 96.97\% & 97.36\% & 90.16\% & 93.74\% \\
      & 34-38 & 11 & 97.98\% & 98.54\% & 91.13\% & 99.55\% & 81.81\% & 92.74\% \\
      & >39 & 9 & 95.76\% & 99.73\% & \textbf{75.51\%} & 94.05\% & 83.47\% & 91.30\% \\
    \hline
    \multirow{4}{*}{\shortstack{Working}} & Full-time work & 48 & - & - & - & - & - & - \\
      & Part-time work & 8 & 95.80\% & 96.44\% & 82.58\% & 85.24\% & 99.67\% & 94.81\% \\
      & Full-time student & 26 & 98.09\% & 99.23\% & 97.83\% & 88.16\% & 85.77\% & 88.37\% \\
      & Others & 8 & 95.80\% & 99.33\% & 98.70\% & 93.59\% & 95.36\% & 99.07\% \\
    \hline
    \end{tabular}
    \caption{Group fairness scores (\textit{GFS}) of three models with different demographic attributes. \textit{GFS} $\geq 80\%$ indicates the fairly treating the minority subgroup; \textit{GFS} $<80\%$ indicates the unfairly treating.}
    \label{tab:GF}
\end{table*}

\section{Discussion}\label{sec:discussion}
In this section, we describe the limitations and implications of our work and discuss possible future directions. 

\subsection{Limitation}

Despite our efforts, the presented work also has its limitations. Firstly, the collected mobility dataset is often biased as they only present a subset of the population who took part in data collection. In many cases, the users are limited to students or those affiliated with the research team that has collected the dataset. This limitation means the examined trajectories are not representative of everyone's mobility behaviour. Furthermore, the demographics of the participants are also limited in terms of age and socio-economic diversity. 

Secondly, in our paper, we reported that we did {\bf not} observe any violation of \emph{group fairness} across gender, age and employment level for the examined PUT models.  However, we acknowledge that the results presented regarding group fairness are highly influenced by the city and societal structures in which the data was collected. In the case of MDC, users' traces correspond to a level of socio-economic and cultural freedom associated with life in Switzerland. Such observations will indeed differ if we examine other cultures, such as those in the United States or Asian countries, where there is a broader socio-economic and gender inequality gap.    
We also believe the availability of datasets with rich demographic information could enable future work to examine the intersection of individual fairness within demographic groups.  Finally, it is worth noting that, unlike online datasets, offline mobility datasets come in limited size due to the great burden the data collection imposes on participants and are handful. Although this limitation could impact the generalization of our results (e.g., that is we cannot claim that Mo-PAE is always fairer than TrajGan), the methods proposed in this study are generalizable and applicable to other PUT models and across mobility datasets. Indeed, we believe future work would focus on creating a toolkit for computing spatial-temporal fairness of datasets and models. We expand on the implications of our work next.

\subsection{Implication}

Our paper has multiple important implications: first, our work offers a novel methodology for defining fairness in the context of spatial-temporal datasets. We believe works such as ours will help shape the future roadmap of Fair-ML studies by offering possibilities to measure equity within different systems such as those of mobility based ones (e.g.,  transportation). The choice of which of the proposed similarity metrics to select for evaluating individual fairness is another critical dimension that could be highly context and application dependent. For example, for applications where there is a need for strict fairness measurement, corresponding to the WYZIWIG worldview~\cite{friedler2021possibility}, a strict similarity measure such as combined entropy (EOTs) could be chosen. In contrast, for applications where the groups are not necessarily equal, but for the purposes of the decision-making process, we would prefer to treat them as if they were, a less sensitive similarity measure such as coarse grain SSIM could be used.   

Although our focus in this work was on fairness analysis of the PUT models, we believe our study can be the first step towards implementing fairness interventions embedded in these models. For example, in-processing approaches rely on adjusting the model during the training to enforce fairness goals to be met and optimized in the same manner as accuracy. This goal is often achieved through adversarial networks or fair representation learning approaches such as~\cite{hu2020fairnn}, model induction, model selection, and regularization~\cite{yan2020fairness}. Of course, designing such mitigation strategies requires access to the underlying architecture of the PUT models which is most of the time not possible, and is in contrast to taking these models as black-box as we did in this study. 

Regarding the relationship between privacy versus fairness, location privacy-preserving mechanisms generally prevent information leakage against protected attributes, and these attributes are also essential to fairness analysis, they are used to ensure little discrimination against protected population subgroups. This dimension also explains why the PUT models achieve group fairness but not individual fairness, as these sensitive attributes considered by group fairness are in protection. The competing trend between individual- and group- fairness also implies another interesting trade-off in Fair-ML.
From the individual perspective, the re-identification risk and individual fairness are in tension. We believe designing privacy-preserving models to become fairness-aware is a research direction that will receive significant attention in the future. 

\section{Conclusion}\label{sec:conclusion}

Intuitively, fairness has a close relationship to privacy, no matter structural data or unstructured data in machine learning. But the quantification between them is still unclear. In this paper, we proposed different metrics for measuring individual fairness in the context of spatial-temporal mobility data. We compared different location privacy-protection mechanisms (PUT models) on the defined individual- and group-based metrics. Our results on two real trajectory datasets show that the privacy-aware models {\bf achieve} fairness at the group level but {\bf violate} individual fairness. Our findings raise questions regarding the equity of the privacy-preserving models when individuals with similar trajectories receive a very different level of privacy gain. We leverage the empirical results of our work to make valuable suggestions for the further integration of fairness objectives into the PUT models. 
Especially when discussing the individual perspective, the tension between the user re-identification task and individual fairness needs to be considered for future spatial-temporal data analysis and modelling to achieve a privacy-preserving fairness-aware setting.




\bibliographystyle{plain}
\bibliography{main-base}
\newpage
\section*{Appendix}

\begin{figure*}
     \centering
     \begin{subfigure}[t]{0.8\textwidth}
     \centering
         \includegraphics[width=\textwidth]{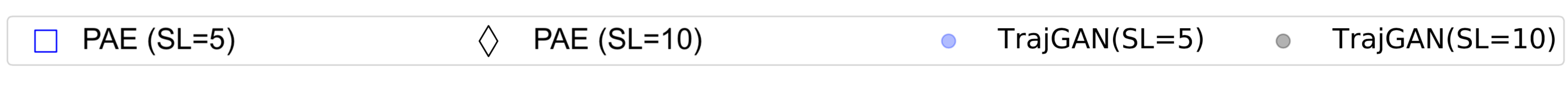}
     \end{subfigure}
     \hfill
     \begin{subfigure}[t]{0.35\textwidth}
         \centering
         \includegraphics[width=\textwidth]{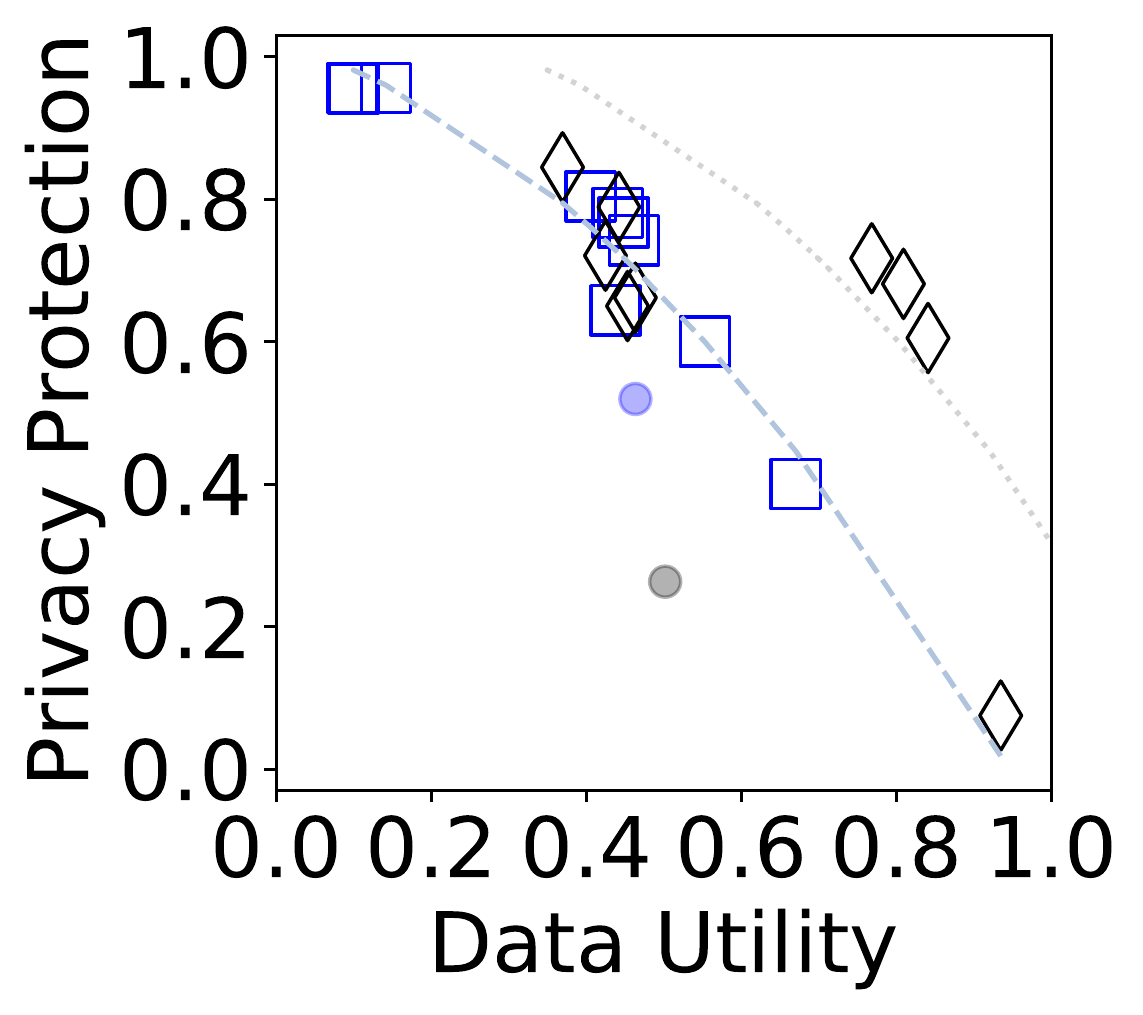}
         \caption{MDC}
         \label{fig:mdc}
     \end{subfigure}
     \begin{subfigure}[t]{0.35\textwidth}
         \centering
         \includegraphics[width=\textwidth]{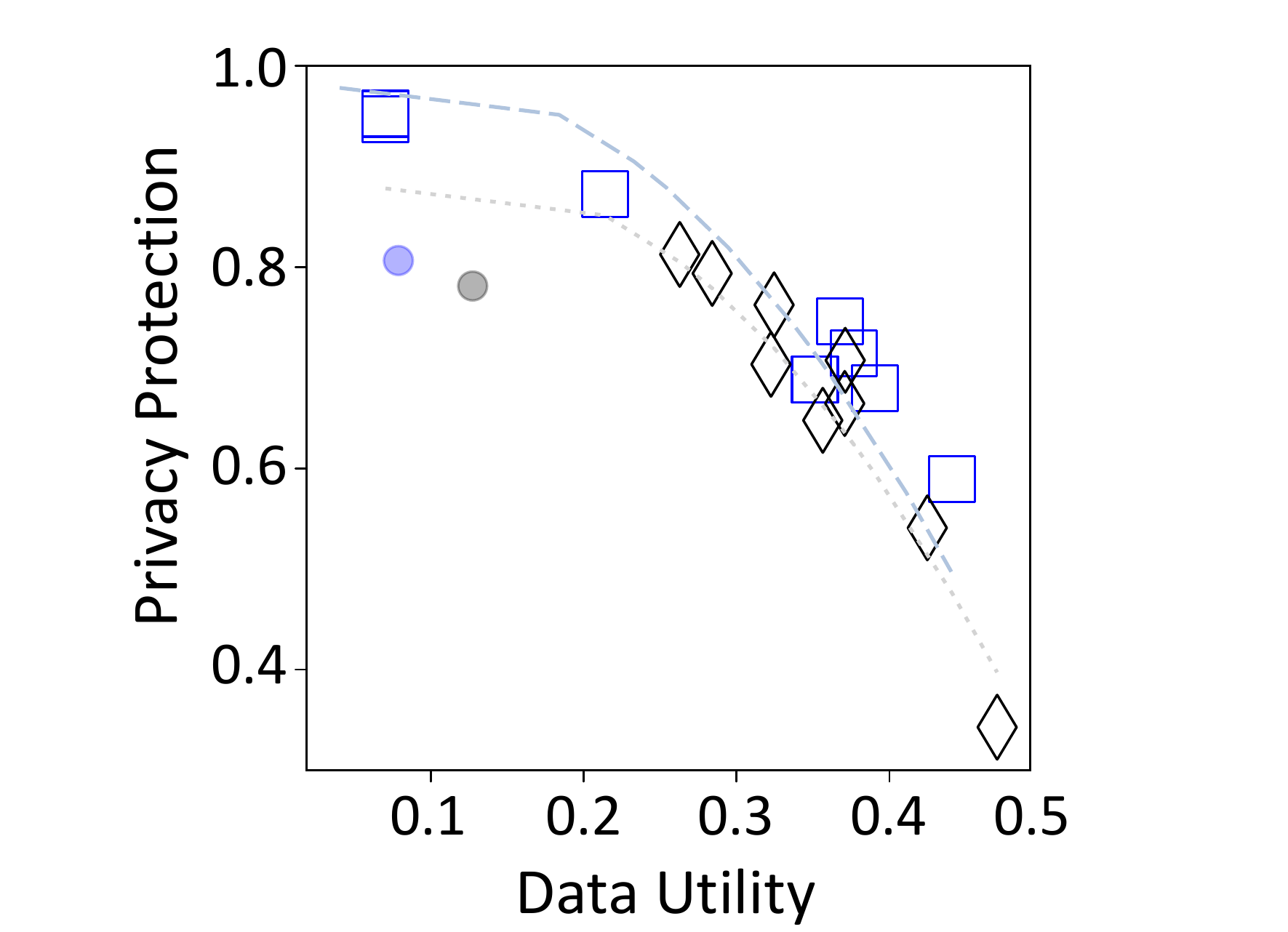}
         \caption{Geolife}
         \label{fig:geolife}
    \end{subfigure}
    \hfill
    \caption{Pareto Frontier trade-off of Utility and Privacy on two datasets. The hollow squares and diamonds present the results of the Mo-PAE models. The solid points present the results of the TrajGAN. Blue color presents sequence length \textit{SL} = 5. Black color presents \textit{SL} = 10.}
    \Description{}
    \label{fig:compare}
\end{figure*}

\subsection{Inference tasks}
Here we list some basic definitions of inference tasks in mobility literature.

\subsubsection{User Re-identification Task (UR)}

The accuracy of the user \linebreak re-identification task is leveraged to assess the \textit{trajectory} \textit{uniqueness} of the mobility trajectory.
With more and more intelligent devices and sensors being utilized to collect information about human activities, the trajectories also expose increasing intimate details about users' lives, from their social life to their preferences. 
A mobility privacy study conducted by De Montjoye et al~\cite{de2013unique} illustrates that four spatial-temporal points are enough to identify 95\% of the individuals in a certain granularity. 
As human mobility traces are highly unique, a mechanism capable of reducing the user re-identification risk can offer enhanced privacy protection in mobility data sharing.
The enhanced privacy protection is referred to \textit{privacy gain} (or \textit{PG}) in the PUT models.

\subsubsection{Mobility Prediction Task \textup{(}MP\textup{)}}

The accuracy of the mobility prediction task is leveraged to assess the \textit{predictability} of the mobility trajectory.
Mobility datasets are of great value for understanding human behaviour patterns, smart transportation, urban planning, public health issue, pandemic management, etc.
Many of these applications rely on the next location forecasting of individuals, which in the broader context can provide an accurate portrayal of citizens' mobility over time.
For the mobility prediction task in this work, the raw geolocated data or other mobility data commonly contain three elements: user identifier \textit{u}, timestamps \textit{t}, location identifiers \textit{l}.
Hence, each location records \textit{r} could be denoted as \emph{$r_i$} = [\textit{$u_i$}, \textit{$t_i$}, \textit{$l_i$}], while each location sequence \emph{S} is a set of ordered location records \emph{$S_n$} =~\{\textit{$r_1$, $r_2$, $r_3$, $\cdots$, $r_n$}\}, namely \textit{mobility trajectory}. Therefore, given the past mobility trajectory \emph{$S_n$} =~\{\textit{$r_1$, $r_2$, $r_3$, $\cdots$, $r_n$}\}, the mobility prediction task is to infer the most likely location \emph{l$_{n+1}$} at the next timestamp \emph{t$_{n+1}$}.
The results of two PUT models indicate that a bit of mobility prediction accuracy is sacrificed in exchange for higher privacy protection. The sacrificed prediction accuracy is referred to \textit{utility decline} in the PUT models.

\subsection{Performance of the Privacy-Utility Trade-off Models}
\label{section:put}

Before examining fairness, we first offer analysis and comparison of the two described PUT models that we are investigating in terms of privacy and utility. Figure~\ref{fig:compare} presents the privacy utility trade-off of Mo-PAE and TrajGAN over the two described datasets. The y-axis presents the privacy gain brought to the raw dataset by applying these models, whereas the x-axis presents the decline in privacy prediction due to this privacy gain. 
The data fed into the Mo-PAE~\cite{zhan2022privacyaware} are a list of trajectories with specific sequence length \emph{SL}, that is \{$S_{sl}^1$, $S_{sl}^2$, $S_{sl}^3$, $\cdots$, $S_{sl}^j$\}. 
For instance, if the sequence length is 10, that indicates each trajectory contains 10 history location records \textit{r}, \emph{$S_{10}$} = \{$r_1$, $r_2$, $r_3$, $\cdots$, $r_{10}$\}, and $SL=10$.

As Mo-PAE is highly dependent on the sequence length and Lagrange multipliers that indicate to what extent privacy or utility must be optimized, each point on the corresponding plots presents experiments with one set of hyper-parameters. These results show that as the Mo-PAE achieves maximum privacy protection it comes with the cost of degrading the prediction accuracy. Similarly, TrajGAN achieves 80\% privacy gain when applied on Geolife Dataset but it highly degrades the utility.  
For the Lagrange multipliers setting of the Mo-PAE in this work, we choose $\lambda_1 = -0.1$, $\lambda_2 = 0.8$, $\lambda_3 = -0.1$, as this combination exerts the most promising privacy-utility trade-off in the Mo-PAE model.

\subsection*{Related Equations}
\label{section:put}
\subsubsection*{i. SSIM}
In this work, we use the known SSIM measure as the perceptual difference of two similar users' heatmap images, $H_i$ and $H_j$:
\begin{equation}
\begin{aligned}
    SSIM(H_i, & H_j) =  \frac{(2\mu_i\mu_j+c_1)(2\sigma_{ij}+c_2)}{(\mu_i^2+\mu_j^2+c_1)(\sigma_{i}^2+\sigma_{j}^2+c_2)}, \\&
    c_1=(k_1L)^2,\ c_2=(k_2L)^2
\end{aligned}
\label{equ:ssim}
\end{equation}

where $\mu_i$ and $\mu_j$ are the averages, $\sigma_i$ and $\sigma_j$ are the variances, and $\sigma_{ij}$ is the covariance of $H_i$ and $H_j$; L is the dynamic range of the pixel-values, $k_1=0.01$ and $k_2=0.03$ by default. 

\subsubsection*{ii. Shannon Entropy (SE)} 
SE is the entropy of the probabilities of visited location distribution. To be specific, this entropy is defined by following the notion in~\cite{wang2020entropy,lu2013approaching} and measured as:

\begin{equation}
E_h= -\sum_{i=1}^n P(x_i)\log_2[P(x_i)]
\end{equation}
where $n$ is the length of the probability vector, $P(x_i$) is the probability of location $x_i$.

\subsubsection*{iii. LonLat Entropy (LE)}
LE is the entropy of the geo-located locations in a time-series format.
This entropy reflects the probability of a new sub-string and quantifies the irregularity or complexity of the time-series data. The $E_f$ of visited longitudes and latitudes are integrated as the LE:

\begin{equation}
E_f= \ln \Phi^m(r,n) - \ln \Phi^{m+1}(r,n)
\end{equation}
where details and default values of the threshold $r$ and the definition of function $\Phi^m(r,n)$ can be found in the study \cite{chen2007characterization,flood2021entropyhub}. 

\subsubsection*{iv. Heatmap Entropy (HE)} 
HE is the entropy of the users' heatmap images. The entropy of trajectory heatmap images was calculated using the two-dimensional sample entropy method ($SampEn_{2D}$) \citep{silva2016two}. In a trajectory heatmap image ($L^2$), the image features were extracted by accounting for the spatial distribution of pixels in different $m$-length square windows with origin at $u(i,j)$.

\begin{equation}
\begin{aligned}
 &E_{2D}\left(u,m,r\right) = -\ln\frac{U^{m+1}(r)}{U^{m}(r)} ,
 \\&U^{m}(r) = \frac{1}{N_m}\sum\nolimits_{i,j,a,b=1}^{i,j,a,b=L-m} Z ,
 \\ & Z = {P\left[x_{m}(a,b)|d\left[x_m(i,j),x_m(a.b)\right] \leq r, (a,b)\neq(i,j)\right]}
\end{aligned}
\end{equation}

where $r$ is the similarity threshold, $N_m$ is the total number of square windows, $P$ is the probability of pixels set $x(i,j)$ satisfying specific conditions, $U_m(r)$ is the average probability, and $d$ is a distance function to calculate the difference of corresponding points.

\subsubsection*{v. Actual Entropy (AE)} 
AE is the entropy of capturing entire spatial-temporal order present in user's mobility pattern.
In this work, the given area is segmented using structured grids, where each grid is initialized as $0$. Then the visited locations and whether the person reached the cell previously are tracked. If the person visits an unreached cell, the location is marked as $1$, generating time-series binary data to characterize the trajectory. The actual entropy $E_a$ is calculated using:

\begin{equation}
E_a= \left(\frac{1}{n}\sum_i \Lambda _i\right)^{-1} \ln (n)
\end{equation}
where $\Lambda _i$ is the length of the shortest sub-string starting at position $i$ which does not previously appear from position 1 to $i-1$, and $n$ is the length of the binary trajectory data.


\end{document}